\documentclass[preprint,review,12pt,3p]{elsarticle}

\usepackage{amssymb}

\usepackage{soul}

\usepackage{amsmath,amssymb}

\usepackage{algorithm}
\usepackage[noend]{algpseudocode}
\usepackage{color}

\makeatletter
\def\BState{\State\hskip-\ALG@thistlm}
\makeatother

\newcommand{\new}[1]{{\color{black}#1}}

\journal{Applied Soft Computing}

\begin{document}

\begin{frontmatter}



\title{Adaptive Verifiability-Driven Strategy for Evolutionary Approximation of Arithmetic Circuits}


\author{Milan \v{C}e\v{s}ka, Ji\v{r}\'{i} Maty\'{a}\v{s}, Vojtech Mrazek, Lukas Sekanina, Zdenek Vasicek, Tom\'{a}\v{s} Vojnar}

\address{Faculty of Information Technology, Brno University of Technology, Czech Republic}

\begin{abstract}
We present a novel approach for designing complex approximate arithmetic circuits that trade correctness for  power consumption and play important role in many energy-aware applications. Our approach integrates in a unique way formal methods providing  formal guarantees on the approximation error into an evolutionary circuit optimisation algorithm. The key idea is to employ a novel adaptive search strategy that drives the evolution towards promptly verifiable approximate circuits. As demonstrated in an extensive experimental evaluation including several structurally different arithmetic circuits and target precisions, the search strategy provides superior scalability and versatility  with respect to various approximation scenarios. Our approach significantly improves capabilities of the existing methods and paves a~way towards an automated design process of provably-correct circuit approximations.
\end{abstract}


\end{frontmatter}



\section{Introduction}

\emph{Approximate circuits} are digital circuits that trade functional
correctness (precision of computation) for various other design objectives such
as chip area, performance, or power consumption. Methods allowing one to develop such circuits are currently in high demand as
many applications require low-power circuits, and approximate circuits---trading
correctness for power consumption---offer a viable solution. Prominent examples of such applications include image and video processing~\cite{median-filters, Vasicek:DATE17}, or architectures for neural networks ~\cite{bio-inspired, Mrazek:iccad16}.

There exists a vast body of literature (see e.g.~\cite{vasicek:sekanina:tec,Nepal:17,mrazek:date:17,Grater:2016})
demonstrating that evolutionary-based algorithms are able  to automatically design innovative
implementations of approximate circuits  providing high-quality trade-offs among
the different design objectives.
%
As shown in~\cite{Ciesielski16, Chand16}, many applications favour \emph{provable error bounds} on resulting 
approximate circuits, which makes automated design of such circuits a
very challenging~task.

To be able to provide bounds on the approximation error, one can, in theory,
simulate the circuit on all possible inputs.
Unfortunately, such an approach does not scale beyond circuits with more than
12-bit operands even when exploiting modern computing
architectures~\cite{Mrazek:iccad16}.
A similar scalability problem does, in fact, emerge already when using
evolutionary optimisation of circuits while preserving their precise
functionality.
To solve the problem in that case, applications of \emph{formal verification
methods}
\cite{vasicek:sekanina:genp:2011,Ciesielski15,Grobner-DATE'16}
have been proposed.
Naturally, attempts to use formal verification methods --- including binary
decision diagrams (BDDs)~\cite{Vasicek:DATE17}, boolean satisfiability (SAT) solving~\cite{MACACO}, model checking~\cite{chand:dac16}, or symbolic computer
algebra employing Gr\"{o}bner bases~\cite{drechsler18} --- have appeared in
design of approximate circuits too.
However, these approaches did still not scale beyond approximation of
multipliers with 8-bit operands and adders with 16-bit operands. 

In this paper, based on our preliminary work~\cite{iccad17}, we propose 
a new approximation technique that integrates formal methods, namely SAT solving, 
into evolutionary approximation.
We concentrate on using \emph{Cartesian genetic programming}
(CGP) for circuit approximation under the \emph{worst case absolute error}
(WCAE) metric, which is one of the most commonly used error metrics.
%
%
The key distinguishing idea of our approach is simple, but it makes our approach
dramatically more scalable comparing to previous approaches.
Namely, we \emph{restrict the resources} (running time) available to the SAT
solver when evaluating a candidate solution.
If no decision is made within the limit, a minimal score is assigned to the
candidate circuit.

This approach leads to a \emph{verifiability-driven search
strategy} that drives the search towards \emph{promptly verifiable approximate circuits.}
%
Shortening of the evaluation time allows our strategy to increase the number
of candidate designs that can be evaluated within the time given for the
entire CGP run. As shown in~\cite{iccad17}, comparing to existing approximation
techniques, our approach  is able to discover circuits that have much better
trade-offs between the precision and
energy savings.


To mitigate negative effects caused by shortening of the evaluation time,  
we propose in this paper  an \emph{adaptive control procedure} that
dynamically adapts the limit on resources available to the SAT solver during the
evolution.
It allows the verification procedure to use more time when needed (typically at
the end of the evolution) in order to discover solutions requiring a longer verification time
and that would be rejected with a fixed resource limit.
On the other hand, the verification time can also be shortened (typically,
though not only, at the beginning of the evolution) when many suitable
candidate designs are produced.

We have implemented the adaptive strategy in ADAC~\cite{ADAC}, our tool for automated design of approximate circuits,
that is now able to discover  complex arithmetic circuits such
as 32-bit approximate multipliers, 32-bit approximate multiply-and-accumulate
(MAC) circuits, and 24-bit dividers providing high quality   trade-offs between the
approximation error and energy savings.
Such circuits have been approximated by a~fully-automated approach with
guaranteed error bounds for the first~time.

\subsection{Contribution}

The main contributions of this work can be summarized as follows:\begin{itemize}

\item We propose a new approach for automated design of approximate arithmetic
circuits that integrates in a unique way formal verification methods for circuit verification into an evolutionary-driven circuit optimisation.

\item We  propose a novel adaptive strategy that controls the evolutionary
search  by introducing adaptive limits on the resources used by the verification
procedure. 

\item Using a detailed experimental evaluation, we demonstrate that the proposed
approach provides high-quality results
for a wide class of approximation problems including circuits with different bit-widths, internal structure, and required precision.
The experiments also show that our approach provides superior scalability and versatility  comparing  to existing approximation methods.

 
\end{itemize}    

Note that while the idea of verifiability-driven search has appeared already in
our preliminary work~\cite{iccad17}, the current paper significantly extends
this work in the following two aspects:
First, we propose and implement the adaptive search scheme that considerably
improves the original verifiability-driven strategy.
\new{In particular, it improves the overall performance, and, more importantly, it
ensures that our approach is versatile, i.e., in contrast to the method described in~\cite{iccad17}, it works well for a wide range of arithmetic circuits and approximation scenarios without manual tuning of the parameters of the evolutionary algorithm. The adaptivity is an important methodological improvement as the versatility is indeed essential for applying the approximation process into automated circuit~design.}

Second, we significantly extend the experimental evaluation to demonstrate the
impact of the features described above.
The evaluation newly includes approximate circuits (with different bit-widths)
for multiplier--accu\-mu\-lators and dividers representing structurally more complex
circuits when compared to adders and multipliers typically used in the
literature. Note that especially MACs play an important role in many energy-aware applications -- for example, MACs represent highly energy  
demanding components  in neural network hardware architectures~\cite{tpu}.

\section{State of the Art}

Various approaches have been proposed to address the problem of rapidly growing
energy consumption of modern computer systems. As one of the most promising
energy-efficient computing paradigms, approximate computing has been
introduced~\cite{Mittal:2016}. Approximate computing intentionally introduces
errors into the computing process in order to improve its energy-efficiency.
This technique targets especially the applications featuring an intrinsic
error-resilience property where significant energy savings can be achieved.  The
inherent error resilience means that it is not always necessary to implement
precise and usually area-expensive circuits. Instead, much simpler approximate
circuits may be used to solve a given problem without any significant
degradation in the output quality. Multimedia signal processing and machine
learning represent typical examples that allow quality to be traded for power,
but approximate computing is not limited to those applications only. A detailed
study of Chippa et al. reported that more than 83\,\% of runtime is spent in
computations that can be approximated~\cite{Chippa:dac2013}.

Many fundamentally different approaches have recently been introduced under the
term of approximate computing. 
%
%
The literature on the subject covers the whole computing stack, integrating
areas of microelectronics, circuits, components, architectures, networks,
operating systems, compilers, and applications. Approximations are conducted for
embedded systems, ordinary computers, graphics processing units, and even
field-programmable gate arrays. A good survey of existing techniques can be
found, for example, in \cite{Mittal:2016,Xu:2016}.

\new{This paper is concerned with automated methods for functional approximation of arithmetic circuits, where the original circuit is replaced by a less complex one which exhibits some errors but improves non-functional circuit parameters. In the following subsections, we provide a brief survey of existing methods for functional circuit approximation. Our approach falls into the category of search-based methods described in Section~\ref{sec:rel-search}. The performance of these methods is directly affected by the performance of the candidate circuits evaluation.  Therefore, in Section~\ref{sec:rel-eval}, we survey existing methods for evaluating the error of approximate circuits.} 

\subsection{Functional approximation}

Technology-independent functional approximation is the most preferred approach
to approximation of digital circuits described at the gate or register-transfer level
(RTL). The idea of functional approximation is to implement a slightly different
function to the original one provided that the accuracy is kept at a desired
level and the power consumption or other electrical parameters are reduced
adequately. The goal is to replace the original accurate circuit (further
denoted as the \emph{golden circuit}) by a less complex circuit which exhibits
some errors but improves non-functional circuit parameters such as power, delay,
or area on a chip. Functional approximation is inherently a multi-objective
optimisation problem with several (typically conflicting) criteria.
%

Functional approximation can be performed manually, but the current trend is to
develop fully automated functional approximation methods that can be integrated
into computer-aided design tools for digital circuits. The fully-automated
methods typically employ various heuristics to identify circuit parts suitable
for approximation.

The Systematic methodology for Automatic Logic Synthesis of Approximate circuits
(SALSA) is one of the first approaches that address the problem of approximate
synthesis~\cite{SALSA}. The authors mapped the problem of approximate synthesis
into an equivalent problem of traditional logic synthesis: the ``don't
care''-based optimisation. 
Another systematic approach,
Substitute-And-SIMplIfy (SASIMI), tries to identify signal pairs in the circuit
that exhibit the same value with a high probability, and substitutes one for the
other~\cite{SASIMI:Venkataramani:date2013}. These substitutions introduce
functional approximations. Unused logic can be eliminated from the circuit,
which results in area and power savings. A different approach was proposed by Lingamneni et al that employed a
probabilistic pruning, a design technique that is based on removing circuit
blocks and their associated wires to trade exactness of computation against
power, area, and delay saving~\cite{Lingamneni:2011}. 

\subsection{Search-based functional approximation}
\label{sec:rel-search}

The main limitation of the techniques based on a variant of probabilistic or
deterministic pruning is the inability to generate novel circuit structures.
None of them allows one to replace a part of the original circuit with a
sub-circuit that does not form a part of the original circuit. This limitation
considerably restricts the space of the possible solutions as shown
in~\cite{mrazek:date:17}. In order to address this issue and improve the quality
of the obtained approximate circuits, various artificial intelligence techniques
have been applied to accomplish approximations. 
Nepal et al. introduced a technique for automated behavioral synthesis of
approximate computing circuits (ABACUS)~\cite{Nepal:17}. ABACUS uses a simple
greedy search algorithm to modify the abstract syntax tree created from the
input behavioral description. In order to approximate gate-level digital
circuits, Sekanina and Vasicek employed a variant of CGP~\cite{vasicek:sekanina:ddecs2014,vasicek:sekanina:tec}. As
shown in~\cite{mrazek:date:17}, this approach is able to produce high-quality
approximate circuits that are unreachable by traditional approximate techniques.
A~comprehensive library of 8-bit adders and multipliers was built using
multi-objective CGP. In the context of FPGAs, circuit approximation has been
introduced and evaluated by means of the GRATER tool~\cite{Grater:2016}. GRATER
uses a genetic algorithm to determine the precision of variables within an
OpenCL kernel. 
%

The proposed search-based approaches share a common idea---they map the problem
of approximate synthesis to a~search-based design problem. An automated circuit
approximation procedure is seen as a multi-objective search process in which a
circuit satisfying user-defined constraints describing the desired trade-off
between the quality and other electrical parameters is sought within the space
of all possible implementations. The approximation process typically starts with
a fully-functional circuit and a target error. A heuristic procedure (e.g.  an
evolutionary algorithm) then gradually modifies the original circuit.
%
%
The modification can affect either the node function (e.g. an AND node can be
modified to an inverter or vice versa), node input connection, or primary output
connection. It is thus able to not only disconnect gates but also to introduce
new gates (by activating redundant gates).
%

\subsection{Evaluating the error of approximate circuits}
\label{sec:rel-eval}

The success of approximate design methods depends on many aspects. Among others,
the efficiency and accuracy of the procedure evaluating the quality of candidate
approximate circuits generated by a chosen heuristic procedure has a~substantial
impact on the overall efficiency. The quality of approximate circuits is
typically expressed using one or several error metrics such as error
probability, average-case error, or worst-case error. 

The search-based synthesis is, in general, computationally expensive (hundred
thousands of iterations are typically evaluated). Hence, the evaluation needs to
be fast as it has a great impact on the scalability of the whole design process.
In order to maintain reasonable scalability and avoid a computationally
expensive exhaustive simulation, many authors simplify the problem and evaluate
the quality of approximate circuits by applying a subset of all possible input
vectors. Monte Carlo simulation is typically utilized to measure the error of
the output vectors with respect to the original
solution~\cite{SASIMI:Venkataramani:date2013,Nepal:17,Jiang:2015}.
Unfortunately, a small fraction of the total number of all possible inputs
vectors is typically used. For example, $10^3$ vectors were used to evaluate a
perceptron classifier and less than $10^4$ vectors were employed for a 16x16
block matcher in~\cite{Nepal:17}; $10^8$ vectors were used to evaluate 16-bit
adders in~\cite{Jiang:2015}. It is clear that this approach cannot provide any
guarantee on the error and makes it difficult to predict the behavior of the
approximate circuit under different conditions. Not only that the obtained error
value strongly depends on the chosen vectors but this approach may also lead to
overfitting. Alternatively, the circuit error can be calculated using a
statistical model constructed for elementary circuit components and their
compositions~\cite{Li:dac2015, Mazahir:tc16}. However, reliable and general
statistical models can only be constructed in some specific situations.

Recently, various applications of formal methods have been intensively
studied in order to improve the scalability of the design process of approximate
circuits.
As said already in the introduction, this step is motivated by the successful
use of such methods when optimising correct circuits (i.e. optimising
non-functional parameters while preserving the original functionality).
%
%
In this area, BDDs have originally been extensively used for combinational
equivalence checking \cite{pixley04}.
Currently, modern SAT solvers are substantially more effective at coping with
large problem instances and large search spaces~\cite{abc-iprove}.
Other successful approaches then include, e.g., symbolic computer algebra based
on Gr\"{o}bner bases \cite{drechsler16-grobner-cec}.

Approaches designed for testing exact equivalence are not directly
applicable for evaluating the approximation error, i.e. for \emph{relaxed
equivalence checking}.
However, the ideas behind efficient testing of exact equivalence can serve as
a~basis for developing efficient methods for checking relaxed
equivalence~\cite{Vasicek:ddecs2017}.
A common approach to error analysis is to construct an auxiliary circuit
referred to as the \emph{approximation miter} \cite{MACACO}. 
This circuit instantiates both the candidate approximate circuit and the golden
circuit and compares their outputs to quantify the error.
The miter is then converted either to the corresponding CNF representation and
further solved using a~SAT solver or represented as a~BDD and analyzed using a
BDD library. 
While SAT solvers are able to handle larger instances, they can be used only
when a binary output is sufficient (typically for the worst case error where one
can ask whether the produced error is under a bound given by the designer as a
parameter~\cite{Mrazek:iccad16}).
On the other hand, BDDs allow one to efficiently examine the set of satisfying
truth assignments which represents a key feature of model counting essential for
calculating average-case error, error probability, or Hamming
distance~\cite{Vasicek:ddecs2017}.
Recently, model checking-based techniques levering the approximate miter~\cite{chand:dac16} as well as 
symbolic computer algebra has also been applied in evaluating and quantifying  errors
of approximate circuits \cite{drechsler18}.

However, the above mentioned approaches do still have a problem to scale
above multipliers with 12-bit operands and adders with 16-bit operands.
This scalability barrier is overcome in our SAT-based approach that can scale
much further due to its verifiability-driven search strategy combined with an
adaptive control of the resource limits imposed on the SAT solver used.

\section{Problem Formulation}

In this section, we formalise the problem of designing approximate arithmetic
circuits as a \emph{single-objective optimisation problem}. Recall that the aim
of the circuit approximation process is to improve non-functional
characteristics (such as the chip area, energy consumption, or delay) of the
given circuit by introducing an error in the underlying computation. 

There exist several \emph{error metrics}
characterising different types of errors such as the worst-case
 error, the mean  error, or the error rate. \new{In this work, we primarily focus on the worst-case error that is 
essential when guarantees on the worst behaviour of the approximate circuits are required. For arithmetic circuits,
the worst-case behaviour  is typically captured by the
\emph{normalized worst-case absolute error} (WCAE) defined as follows.}

For a golden (original) circuit~$G$, which computes a~function $f_G$, and its
approximation $C$, which computes a~function $f_C$, where $f_G,f_C:\{0,1\}^n
\rightarrow \{0,1\}^m$, 

$$\mbox{WCAE}(G,C)
= \max_{x\in \{0,1\}^n} \frac{\left|\mathrm{int}(f_G(x)) -
\mathrm{int}(f_C(x))\right|}{2^{m}-1}.$$

\new{Alternatively, the worst-case behaviour can be characterised by the worst-case relative error or maximal Hamming distance.
To simplify the presentation of the main contribution of this work, a novel adaptive verifiability-driven approximation, we restrict ourselves to  WCAE. Note that, only the miter construction in SAT-based candidate circuit evaluation (see Section~\ref{sec:circuiteval}) has to be adapted to work with other worst-case error metrics. Moreover, as shown in~\cite{iccad17}, there is a close relation between the circuits optimised for WCAE and for the mean absolute error representing another important metric that requires more complex evaluation procedure.}


Non-functional characteristics of the circuit, such as the delay, power
consumption, or chip area, depend on the target technology the circuit is
synthesised for.
Computing these characteristics precisely for every candidate solution would
introduce a significant computation burden for the approximation process.
Therefore, we approximate these characteristics by an estimated size of the
circuit computed as follows.
We assume that we are given a list of gates that can be used in the circuit and
that each gate is associated with a constant characterising its size.
The size of the particular gates is specified by the users and should respect
the target technology (cf.  Table~\ref{tab:gates} for the gates and their sizes
used in our experiments).
For a~candidate circuit $C$, we then define its size, denoted
$\mathrm{size}(C)$, as the sum of the sizes of the gates used in $C$. 
As shown in~\cite{Mrazek:iccad16,Vasicek:DATE17,image}, $\mathrm{size}(C)$
typically provides a good estimate for the chip area as well as for the power
consumption.

The problem of finding the best trade-offs between the circuit size and the
WCAE, can be naturally seen as a~multi-objective optimisation problem. In our
approach, we, however, treat it as a series of single-objective problems where
we fix the required values of the WCAE. This approach is motivated by the fact
that the WCAE is usually given by the concrete application where the approximate
circuits are deployed. Moreover, as shown in several
studies~\cite{vasicek:so-mo}, optimising the chip size for a fixed error allows
one to achieve significantly better performance compared to more general
multi-objective optimisation producing Pareto fronts. The performance directly
affects the time required to find high-quality approximation and is essential to
scale to complex circuits such as 16-bit multipliers and beyond. 

The key optimisation problem we consider in the paper is formalised as follows:

\vspace{0.5em}\noindent \textbf{Problem:} \emph{For a given golden circuit $G$
and a threshold $\mathcal{T}$, our goal is to find a circuit $C^*$ with the
minimal size such that the error $\mbox{WCAE}(G,C^*) \leq
\mathcal{T}$.}\vspace{0.5em}
 
Before presenting our approach, we emphasise that our aim is not to provide a
complete algorithm that guarantees the optimality of $C^*$: such an algorithm
clearly exists as the number of circuits with a given size is finite, and one
can, in theory, enumerate them one by one. We rather design an effective search
strategy that is able to provide high-quality approximations for complex
arithmetic circuits having thousands of gates in the order of hours.


\section{Adaptive Verifiability-driven Optimisation}

In this section, we propose our novel optimisation scheme employing four key
components: (1) a~\emph{generator} of candidate circuits that builds on
\emph{Cartesian Genetic Programming} (CGP), (2) an \emph{evaluator} that
evaluates the error of the candidates by leveraging SAT-based verification
methods, (3) a~\emph{verifiability-driven search}  integrating the cost of the
circuit evaluation into the fitness function, and (4) an~\emph{adaptive
strategy} adjusting the allowed cost of evaluation of candidate solutions during
the approximation process.  

\subsection{Generating candidate circuits using CGP}

\begin{figure}[h]\centering%
\includegraphics[width=\columnwidth]{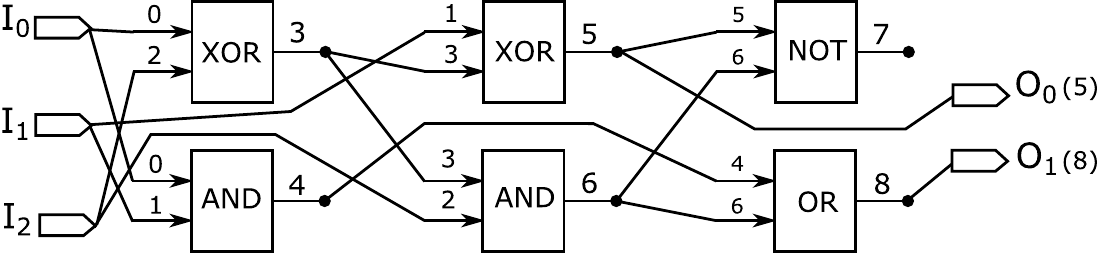}%
\caption{A full adder represented by CGP. Chromosome: (0,~2,~2) (0,~1,~0)
(1,~3,~2) (3,~2,~0) (5,~6,~3) (4,~6,~1) (5,~8), node functions: AND (0), OR (1),
XOR (2), NOT (3).}
\label{fig:cgp-ex}
\end{figure}

CGP is a form of genetic programming where candidate solutions are represented
as a string of integers of a~fixed length that is mapped to a directed acyclic
graph~\cite{miller:cgp:book}. This integer representation is called a
\textit{chromosome}. The chromosome can efficiently represent common
computational structures including mathematical equations, computer programs,
neural networks, and digital circuits. In this framework, candidate circuits are
typically represented in a two-dimensional array of programmable two-input
nodes. 
%
%
The number of primary inputs and outputs is constant. In our case, every node is
encoded by three integers in the chromosome representation where the first two
numbers denote the node's inputs (using the fact that each input of the
circuit and the output of each gate is numbered), and the third represents the
node's function (see the illustration in~Fig.~\ref{fig:cgp-ex}). The codes
of the gates are ordered column-wise. At the end of the chromosome, outputs of
the circuit are encoded using the numbers of gates from which they are taken. The
so-called level-back parameter specifies from how many levels before a given
column the source of data for the gates in that column can be taken.

We use a standard CGP that employs the (1+$\lambda$) search method where a
single generation of candidates consists of the parent and $\lambda$ offspring
candidates. The fitness of each of the solutions is evaluated and the best
solution is preserved as the parent for the next generation. Other candidates
from the generation are discarded.

In circuit approximation, the evolution loop typically starts with
a~\emph{parent} representing a correctly working circuit. New candidate circuits
are obtained from the parent using a \textit{mutation operator} which performs
random changes in the candidate's chromosome in order to obtain a new, possibly
better candidate solution. The mutations can either modify the node
interconnection or functionality. The number of the nodes of candidate circuits
is reduced by making some nodes inactive, i.e. disconnected from the outputs
of the circuit. However, since such nodes are not removed, they can still be
mutated and eventually become active again.



The whole evolution loop is repeated until a termination criterion (in our case,
a time limit fixed for the evolution process) is met. For more details of CGP,
see~\cite{miller:cgp:book}.



\subsection{Candidate circuit evaluation}
\label{sec:circuiteval}

Recall that the candidate circuit evaluation takes into consideration two
attributes of the circuit, namely, whether the approximation error represented
by WCAE is smaller than the given threshold and the size of the circuit.
Formally, we define the fitness function~$f$ in the following way:

$$\begin{aligned} f(C) = \begin{cases} \mathrm{size}(C) & \mbox{ if }
\mbox{WCAE}(G,C) \leq \mathcal{T}, \\ \infty  & \mbox{ otherwise.} \end{cases}
\end{aligned}$$\\[-5mm]

The procedure deciding whether $\mbox{WCAE}(G,C) \leq \mathcal{T}$ represents
the most time consuming part of the design loop. Therefore, we call the
procedure only for those candidates $C$ that satisfy that ${\mathrm{size}(C)
\leq \mathrm{size}(B)}$ where $B$ is the best solution with an acceptable error
that we have found so far. 

\begin{figure}\centering%
\includegraphics[width=.8\columnwidth]{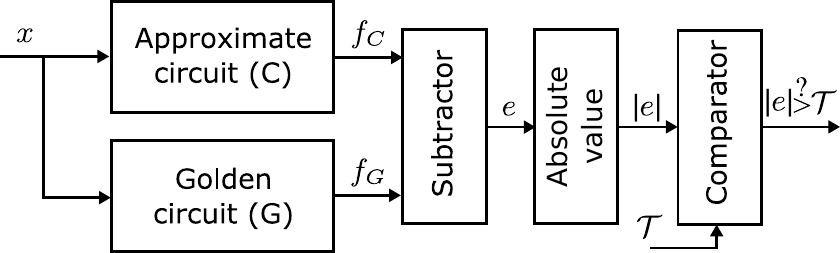}%
\vspace{-5pt}%
\caption{A high-level view on the typical approximation miter for the WCAE
analysis, typically $e(x) =f_G(x)-f_C(x)$.}
\label{fig:miterg}%
\vspace{-10pt}%
\end{figure}

To decide whether $\mbox{WCAE}(G,C) \leq \mathcal{T}$, we adopt the concept of
an \emph{approximation miter} introduced in~\cite{MACACO,Error-Quant-DAC'16}.
The miter is an auxiliary circuit that consists of the inspected approximate
circuit~$C$ and the golden circuit $G$ which serves as the specification. $C$
and $G$ are connected to identical inputs. A subtractor and a comparator then
check whether the error introduced by the approximation is greater than a~given
threshold $\mathcal{T}$. The high-level structure of the approximation miter is
shown in Fig.~\ref{fig:miterg}. The output of the miter is a single bit which
evaluates to logical 1 if and only if the constraint on the WCAE is violated for
the given input $x$.

Once the miter is built, it is translated to a Boolean formula that is
satisfiable if and only if ${\mbox{WCAE}(G,C) > \mathcal{T}}$. This approach
allows one to reduce the decision problem to a~SAT problem and use existing
powerful SAT solvers. Of course, this is a high-level view only. On the
gate-level, we optimize the miter construction by using a novel circuit
implementation of the subtractor, absolute value, and comparator nodes as
described in the conference paper~\cite{iccad17}. The construction, whose
details we skip here since they are rather hardware-oriented, leads to
structurally less complex Boolean formulas. In particular, it avoids long XOR
chains, which are a~known cause of poor performance of the state-of-the-art SAT
solvers~\cite{han:cav12}. For details see~\cite{iccad17}.

\subsection{Verifiability-driven search}\label{section-vdsearch}

During our initial experiments with the approximation of large circuits, we
discovered that the time required for the miter-based circuit evaluation can
significantly differ even among structurally very similar candidates. For
example, there are 16-bit approximate multipliers where checking that
${\mbox{WCAE}(G,C) \leq \mathcal{T}}$ holds takes less than a second, however,
other similar approximations require several minutes. Additionally, we observed
that the more complex the circuits to be approximated are, the higher are the
chances that the evolution stumbles upon a solution that requires a prohibitive
evaluation time. If such a candidate is accepted as a parent, its offspring are
likely to feature the same or even longer evaluation time. Therefore, the whole
evaluation process slows down and does not achieve any significant improvements
in the time limit available for the entire optimisation.

To alleviate this problem, we propose a verifiability-driven search strategy
that uses an additional criterion for the evaluation of the circuit $C$. The
criterion reflects the ability of the decision procedure, in our case a SAT
solver, to prove that ${\mbox{WCAE}(G,C) \leq \mathcal{T}}$ with a given limit
$L$ on the resources available. It leverages the observation that a long
sequence of candidate circuits $B_i$ improving the size and having an acceptable
error has to be typically explored to obtain a~solution that is sufficiently
close to an optimal approximation $C^*$. Therefore, both the SAT and the UNSAT
queries to the SAT solver have to be short. If the procedure fails to prove
$\mbox{WCAE}(G,C) \leq \mathcal{T}$ within the limit~$L$, we set $f(C)=\infty$
and generate a new candidate. 


The interpretation of the resource  limit $L$ on checking that $\mbox{WCAE}(G,C)
\leq \mathcal{T}$ depends on the implementation of the underlying satisfiability
checking procedure. Note that a time limit is not suitable since it does not
reflect how the structural complexity of candidate circuits affects the
performance of the procedure. Therefore, we employ the limit on the
\emph{maximal number of backtracks} in which a single variable can be involved
during the backtracking process (also called the maximal number of \textit{conflicts} on a variable). As the backtracking represents the key and
computationally demanding part of modern SAT solvers~\cite{sat-backtrack}, it
allows one to effectively control the time needed for particular evaluation
queries.  Moreover, it takes into account the structural complexity of the
underlying boolean formula capturing the complexity of the circuit.

\begin{figure}%
    \centering%
    \includegraphics[width=1\linewidth]{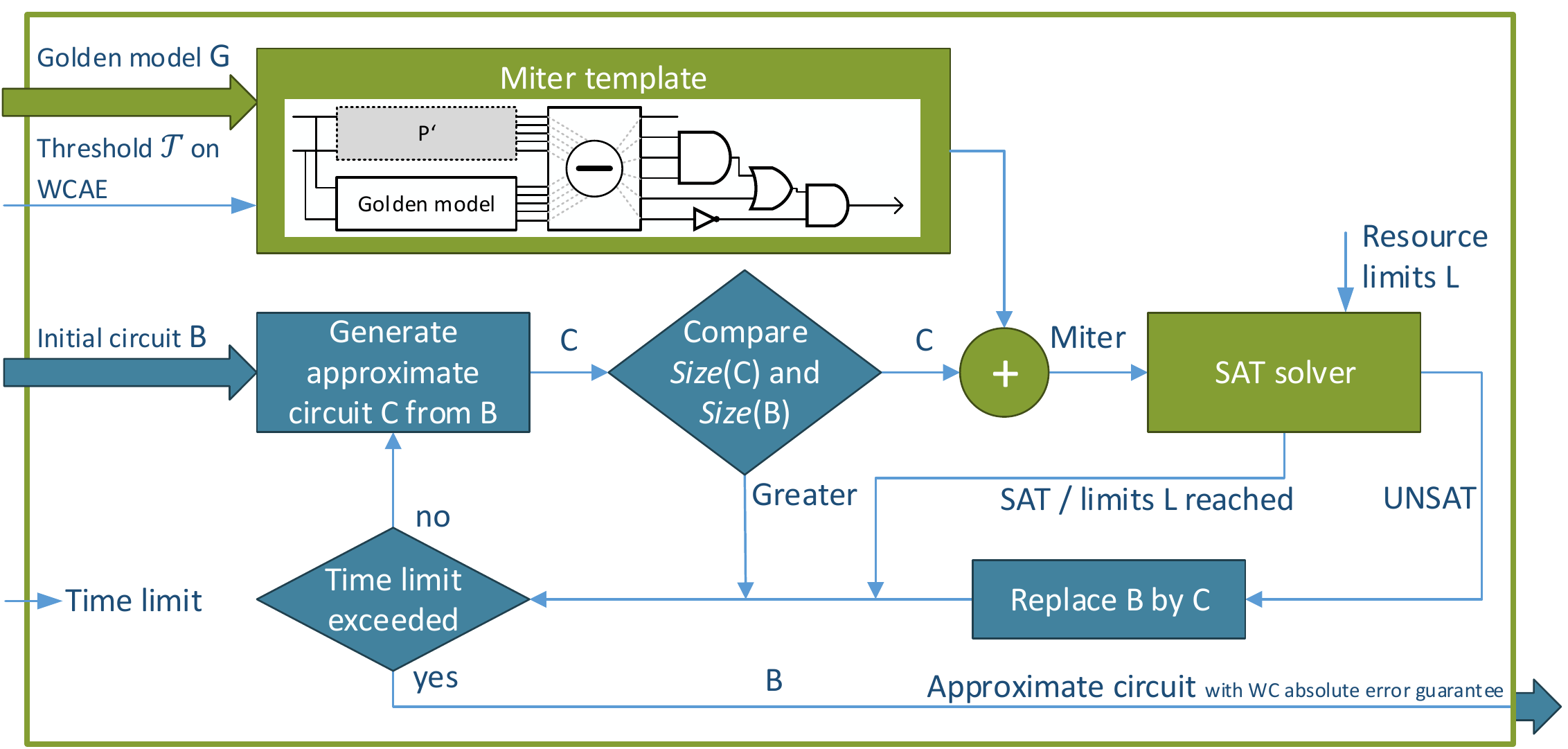}%
    \vspace{-5pt}%
    \caption{The main steps of the proposed verifiability-driven search scheme.
   }\label{fig:cgp}
    \vspace{-15pt}%
\end{figure}

The overall optimisation loop using the verifiability-driven search is
illustrated in~Fig.~\ref{fig:cgp}. The inputs of the design process
include: (1) the golden model~$G$, (2) the threshold on the worst-case absolute
error~$\mathcal{T}$, (3) the initial circuit $B$ having an acceptable error (it
can be either the golden model or its suitable approximation that we want to
start with), and (4)~the time limit on the overall design process. The loop
exploits the CGP principles for the case of $\lambda = 1$, i.e. for
populations consisting of the parent and a single child, which turns out to be a
suitable setting in our experiments discussed below. In other words, the loop
uses mutations to generate a single new candidate circuit $C$ from the
candidate circuit~$B$ representing the best approximation of the circuit $C^*$
that we have found so far. The circuit~$C$ is then evaluated using the fitness
function $f$ as described above. If the candidate $C$ belongs to an improving
sequence (i.e.~${\mathrm{size}(C) \leq \mathrm{size}(B)}$ and $\mbox{WCAE}(G,C)
\leq \mathcal{T}$), we replace $B$ by $C$. The design loop terminates if the
overall time limit is reached, and $B$ is returned as the output of the design
process.

\subsection{Adaptive resource limit strategy}\label{section-adaptive}

In our original conference paper~\cite{iccad17}, we performed a~preliminary
experimental evaluation of the verifiability-driven search strategy studying how
the limit $L$ on the maximal number of backtracks in the SAT decision procedure
affects the performance of the approximation process applied on multipliers
and adders of various bit-widths. In particular, we considered 20K, 160K, and
unboundedly many backtracks. The results clearly demonstrated that the
evolutionary algorithm found best solutions for the lowest of these three limit
settings for a wide range of circuits. However, the question whether a~still
lower SAT limit would improve the performance even further remained open.
Likewise, there remained a question what limits would be appropriate for
other circuits than those considered in the experiments.

Apparently, the lower the limit is the faster the evaluation of each candidate
solution will be. This results in processing a higher number of generations in a
given time interval, hopefully leading to better results. On the other hand,
aggressive limit settings reduce the search space of candidate solutions that
can be evaluated within the given limit. A too tight restriction might prevent
the candidate solutions from diverting from the original solution and reaching
significant improvements (most of the newly generated candidates will likely be
skipped due to exceeding the evaluation limit). Also, the type and complexity of
the approximated circuit and the approximation error can play a significant role
in choosing ideal limit settings. Thus, to reach the best performance of the
method, each new instance of the problem would require an evaluation of
different limit values. Moreover, a fixed limit value might not be optimal
during the course of the evolutionary process even if it is optimal in some
of its phases.

Therefore we propose a new \emph{adaptive strategy} that alters the limit within
the evolutionary run and tries to set it to the most suitable value with regards
to the recently achieved progress. We designed the strategy scheme based on our
previous observations that the limit should be kept low in the early stages of
the evolution so that the clearly redundant logic can be quickly eliminated.
Later in the evolutionary process, the algorithm converges to a locally optimal
solution and improvements in the fitness cease to occur. When such a stage is
reached, the limit needs to be increased in order to widen the space of feasible
candidate solutions at the expense of slower candidate evaluation. Moreover,
once some more significantly changed solution is found, it may again be possible
to shorten the time limit needed for the evaluation, and the process of
extending and shrinking the time limit may repeat (as witnessed also in our
experiments).

\begin{algorithm}[h]
\caption{Adapting the time limit for evaluating candidates}\label{adaptive-alg}
\begin{algorithmic}[1]
\State $lastGens \gets 0$
\State $improvementCount \gets 0$
\Function{updateLimit}{limit, improvement}
\State $lastGens \gets lastGens + 1$

\If {$improvement$}
    \State {$improvementCount \gets improvementCount + 1$}
\EndIf

\If {$lastGens \text{ mod period} = 0$}
    \If {$improvementCount > \tau_{dec}$}
        \State {$limit \gets limit - \delta * limit$}
    \ElsIf {$improvementCount < \tau_{inc}$}
        \State {$limit \gets limit + \delta * limit$}
    \EndIf
    \State {$improvementCount \gets 0$}
    \State {$lastGens \gets 0$}
\ElsIf {$improvementCount > \tau_{res}$}
    \State {$lastGens \gets 0$}
    \State {$improvementCount \gets 0$}
    \State {$limit \gets limit - \delta * limit$}
\EndIf

\State $limit \gets max(limit, minLimit)$
\State $limit \gets min(limit, maxLimit)$
\State \Return limit
\EndFunction
\end{algorithmic}
\end{algorithm}

Our strategy is
described in pseudocode in Algorithm~\ref{adaptive-alg}. The strategy changes
the limit during the evolution process and is driven by four main parameters and
two additional limit values with the following semantics:\begin{itemize}

  \item $period$: the number of generations after which a \emph{periodic
  check} whether the evaluation limit should be changed is triggered.

  \item $\delta$: the \emph{increase/decrease ratio} which says by what
  fraction of the current limit the limit is increased/decreased when such a
  change is considered useful.

  \item $\tau_{dec}$: if the number of improvements that occur in a period
  is above this threshold, the time limit for the evaluation will be
  \emph{decreased}.
  
  \item $\tau_{inc}$: if the number of improvements that occur in a period
  is below this threshold, the time limit for the evaluation will be
  \emph{increased}.

  \item $\tau_{res}$: if this threshold is hit, an \emph{immediate decrease}
  of the time limit and a reset of the generation counter is triggered. This
  threshold applies when the limit becomes clearly too high, which can happen as
  witnessed by our experiments.

  \item $minLimit$: a \emph{minimum limit bound} that restricts the possible
  values of the time limit achievable by the adaptive strategy from below.

  \item $maxLimit$: a \emph{maximum limit bound} that restricts the possible
  values of the time limit achievable by the adaptive strategy from above.

\end{itemize}

Algorithm~\ref{adaptive-alg} allows the strategy to track the current progress
of the evolutionary algorithm and adapt the resource limit accordingly. The key
purpose of the algorithm is to keep the limit low while the evolutionary process
achieves improvements in the candidate solutions and increase the available
resources once the progress is seemingly stalled by the imposed limit. The
control algorithm tracks the number of improvements made in the last $lastGen$
generations in a global variable $improvementCount$. If the number of current
improvements exceeds the value of $\tau_{res}$, the limit is immediately
decreased. Otherwise, the algorithm waits until the $period$ number of
generations is reached and then either increases or decreases the limit based on
the comparison of the $improvementCount$ and the thresholds $\tau_{inc}$ or
$\tau_{dec}$, respectively.

The value of the increment/decrement of the resource limit is relative to the
current limit value. This allows the strategy to both delicately alter small
limit values and reach high limit values in reasonable time. The limit value is
restricted to stay within the interval $\langle minLimit, maxLimit \rangle$.
This ensures that we do not get too small limit values that would reject all
candidates nor too big limit values that would feature a~very long evaluation
time, which would practically stop the approximation process.


\section{Experimental Evaluation}\label{sections-experiments}

In this section, we present a detailed  experimental evaluation of the proposed
method for evolutionary-driven circuit approximation. We first describe the
experimental setting and briefly discuss the CGP parameters we used in the
evaluation. Afterwards, we present a thorough evaluation of the adaptive feature
of our approach as well as an overall comparison of our approach with other
existing approaches. In particular, our experiments focus on answering the
following research questions:
\begin{itemize}

  \item[\textbf{Q1}] Can the adaptive strategy reduce the randomness of the
  evolution-based approximation process?

  \item[\textbf{Q2}] \new{Can the adaptive strategy efficiently handle 
  different circuit approximation problems -- is it more versatile  than the fixed-limit strategies?}
  

  \item[\textbf{Q3}] Can the adaptive strategy outperform the best
  fixed-limit strategy for a given circuit approximation problem?

  \item[\textbf{Q4}] Does the proposed method significantly outperform other
  circuit approximation techniques?

\end{itemize} 

\subsection{Experimental setup}

The proposed circuit approximation method was implemented in our tool called
ADAC---Automatic Design of Approximate Circuits~\cite{ADAC}. ADAC is implemented
as a module of ABC~\cite{abc-iprove}, a state-of-the-art academic tool for
hardware synthesis and verification. ABC provides means for exact
equivalence checking but also general SAT solving. We use the latter for
solving our approximation miters. 





In the experiments, we consider the following circuits for evaluating the
performance of the proposed method\footnote{Gate-level implementations of the considered multipliers and MACs were designed
using the Verilog ``$*$'' and ``$+$'' operators and subsequently synthesised by
the Yosys hardware synthesis tool using the gates listed in
Table~\ref{tab:gates}. Gate-level representations of the dividers were created
according to~\cite{algebraic-circuits}.}:

\begin{itemize} \itemsep 4pt

  \item 16-bit multipliers (the input is two 16-bit numbers) having 1525 gates
  (501 xors and logic depth 34),

  \item 24-bit multipliers having 3520 gates (1157 xors and logic depth 40), 

  \item 24-bit multiply-and-accumulate (MAC) circuits (the input is two 12-bit
  numbers and one 24-bit number) having 1023 gates (321 xors and logic
  depth 39),

  \item 32-bit MAC circuits having 1788 gates (565 xors and logic depth 44),

  \item \new{20-bit squares (the input is one 20-bit number, the result is second power of the input) with 2213 gates (789 xors and logic depth 38)},
  
  \item \new{28-bit squares with 4336 gates (1547 xors and depth 40)}.

  \item 23-bit dividers (the input is 23-bit and 12-bit numbers) having 1512
  gates (253 xors and logic depth 455),

  \item 31-bit dividers with 2720 gates (465 xors and depth 799),

\end{itemize}

Recall that we consider the circuit size as the key non-functional characteristic we want to improve by allowing an error in the circuit computation.  To estimate the circuit size, we use the gate sizes listed in
Table~\ref{tab:gates}. These sizes correspond to the 45nm technology which we
consider in Section~\ref{soa-comparison} when comparing the power-delay product\footnote{Power-delay product is a standard characterisation capturing both the circuit power consumption and performance.} of our resulting circuits with state-of-the-art solutions.

\begin{table}[]
\centering
\caption{Sizes of the gates used in the experiments. The sizes are in
$\mu m^2$ and correspond to the 45nm technology.}
\label{tab:gates}
\begin{tabular}{|l|l|l|l|l|l|l|l|}
\hline
\textbf{Gate} & INV & AND & OR & XOR & NAND & NOR & XNOR \\ \hline
\textbf{Size} & 1.40 & 2.34 & 2.34 & 4.69 & 1.87 & 2.34 & 4.69 \\ \hline
\end{tabular}
\end{table}



\vspace{1em}
\noindent
\emph{Justification for the selected benchmark set:} 

Approximation of 16-bit multipliers represents the cutting edge of
circuit approximation techniques due to the circuit size (i.e. the number of
gates) and structural complexity (i.e. the presence of carry chains), especially
when some formal error guarantees are expected from the approximation method. We
use such multipliers in Section~\ref{soa-comparison} to compare our approach
with state-of-the-art techniques. The other circuits we consider go beyond this
edge: MACs have a~more complicated structure and the error of the involved
multiplication is further propagated in the consequent accumulation. \new{Square circuits computing the second power of the input represent a specialised version of multipliers, while these circuits feature less inputs than other examined instances, their internal structure is much more complex than the structure of arithmetic circuits with comparable input bit widths.}
Approximation of dividers represents a true challenge since they are
structurally more complicated, much deeper, and significantly less explored
(e.g. when compared with multipliers). 

\new{For all 8 circuits, we consider various WCAE values, namely, we let WCAE range from $10^{-4} \%$ to $1 \%$.
The given bound on the WCAE value determines permissible changes in the circuit structure (i.e. a small error allows only smaller changes in the circuit). Therefore,  different WCAE values lead to significantly different approximation problems. We also consider two time limits (1 and 6 hours) for the approximation process. Note that, the time limit also considerably affects the approximation strategy as the given time has to be effectively used with respect to the complexity of approximation problem}

\new{In our experimental evaluation, we explore all three dimensions characterising the circuit approximation problems: i) the circuit type reflecting both the size and the structural complexity, ii) the error bound, and iii) approximation time. In total, we examine more than 70 instances of the approximation problems that sufficiently cover practically relevant problems in the area of arithmetic circuits approximation. Therefore, the considered benchmark allow us to answer the research questions and, in particular, to robustly evaluate the versatility of the adaptive strategies and their benefits with respect to the fixed-limit strategies.}

\new{Note that we exclude adders from our  experimental evaluation as they represent a much simpler approximation problem -- comparing to 16-bit multipliers, 128-bit adders have only around 1/3 of the gates and are structurally less complex. Therefore, the miter-based candidate evaluation handle these circuits without leveraging the resource limits.}


\subsection{CGP parameters}

 The performance of CGP for particular application domains can be tuned by
various CGP parameters out of which the following are relevant in our
case:\begin{itemize} \itemsep 4pt

  \item the number of offsprings ($\lambda$),

  \item the frequency of mutations, and

  \item the CGP grid size and the \texttt{L-back} parameter (i.e.
  connectivity in the chromosome).

\end{itemize} We now briefly discuss our choice of the values of these
parameters that will later be used for the main part of the evaluation of the
proposed method. 

The literature shows that, for a fixed number of generated and evaluated
candidate solutions, CGP-based circuit optimization (i.e. when circuits are not evolved from scratch) with a~smaller value of
$\lambda$ usually leads to better fitness values than CGP using larger~values of
$\lambda$~\cite{vasicek:eurogp15}. 
%
%
%

\begin{figure}
\includegraphics[width=\columnwidth]{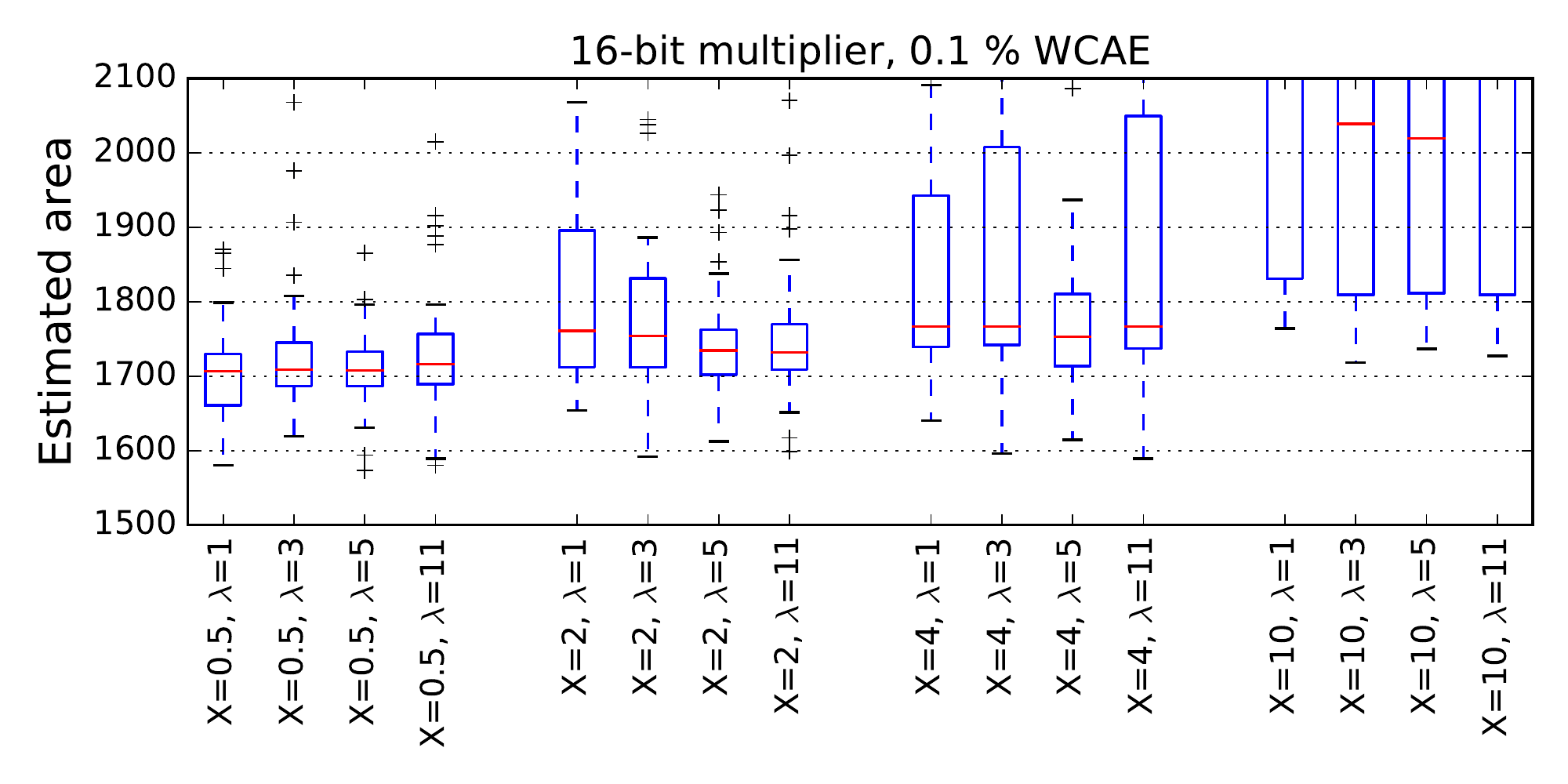}
\vspace{-2em}
  \caption{The impact of the number of offsprings ($\lambda$) and mutation frequency~($X$) on the final
  circuit area of approximated 16-bit multipliers obtained by CGP with a fixed
  time limit for each evolutionary run.}

\label{fig:pop-mut-comparison}
\end{figure}

Aside from the population size, we also examine the effect of the mutation frequency on the performance of circuit approximation. Each time the mutation operator is applied, it alters a~single integer in the
chromosome. When we generate a new candidate from a parent, we apply the
mutation operator up to $M$ times, $M = 0.01 * X * gates(G)$ where $X$ is the
mutation frequency parameter and $gates(G)$ is the number of gates of the
golden solution $G$ that is approximated. In our particular experiment, in which
$gates(G) = 1525$, the performance was evaluated for $X \in \langle
0.5,10\rangle$, i.e. $M \in \langle 8, 153 \rangle$ mutations per chromosome.

\new{
Fig.~\ref{fig:pop-mut-comparison} provides the results of approximation of 16-bit multipliers with 0.1~\% WCAE using different combinations of $\lambda$ and $X$ (the $x$-axis). The $y$-axis characterises the size (obtained as the sum of gate sizes) of the best candidate found in every $2$-hour run. The SAT resource limit was set to $100$. We do not present results for other approximate circuits as they exhibit similar patterns. The boxplots are grouped by mutation frequency. We can see that the performance within each group is very similar and lower mutation frequencies perform better than higher mutation frequencies. We also applied Friedman and Nemenyi statistical tests~\cite{friedman:1940, demsar:2006, pmcmr} to evaluate these results. According to Nemenyi post hoc test, the differences between various $\lambda$ values within the same mutation frequency are not significant at $\alpha = 0.05$. Mutation frequencies $X=0.5$ and $X=2$ are equivalent and perform significantly better than $X=4$ and $X=10$.}

Our experiments confirm general observations known from the literature (see, e.g.,~\cite{Mrazek:iccad16}): the number of mutations should be small. This way,
the mutations perform slight changes between the generations only. Otherwise,
for a high mutation frequency, the function of a new solution is usually
completely altered. Such a solution is then rejected with a high probability,
the search gets close to a random one, and its efficiency deteriorates.
Therefore, in the rest of the experiments, we choose the mutation frequency $X =
0.5~\%$. As population size does not seem to significantly matter, we choose the simplest $\lambda = 1$ scheme.

Finally, we set the dimensions of the chromosome gate matrix as $1 \times W$
where $W$ equals the number of gates in the correct circuit (i.e. all the gates
of the original circuit are in one row) and use \texttt{L-back} = $W$, i.e. we
allow the maximum connectivity of the chromosomes. This setting gives the
evolutionary algorithm maximal freedom with respect to the candidate solutions
that can be created. This fact is desirable in case of area
optimization we aim at~\cite{vasicek:so-mo, miller:cgp:book}.

\subsection{Comparison of adaptive strategies}

\new{In the next phase of our experiments, we examine different versions of  the
adaptive strategy corresponding to different instantiations of the adaptive
scheme presented in Section~\ref{section-adaptive}. The goal of this phase is to
select the best adaptive strategies that efficiently works for a wide class of approximation problems.
These strategies are further thoroughly evaluated and compared with
fix-limit strategies on the selected benchmark.}    

\new{Based on our experience with the limit values used in~\cite{iccad17}, we
consider five versions of the adaptive strategy given by the parameter values
listed in Table~\ref{table-ada-params}. These versions have been chosen to 
adequately cover the space of adaptive strategies and thus they range
from strategies that try to promptly react to changes in the evolutionary
process ($ada1$, $ada3$) to strategies that evaluate the progress of the
evolution over longer periods of time ($ada2$). The remaining strategies
($ada4$, $ada5$) lie in the middle of the range.}

The strategies differ mainly in two basic aspects: the length of the period
with which the adaption happens and the thresholds used for the adaption
($\tau_{dec}$, $\tau_{inc}$, $\tau_{res}$). Larger values of the thresholds
with respect to the $period$ mean that the resource limit will more likely be
increased. Strategies with such thresholds ($ada1$, $ada3$ and $ada5$) are
faster to magnify the limit once the evolution seemingly gets stuck in a local
optimum. Thus, the possible search space is broadened, but each candidate
evaluation is likely to take longer time. On the other hand, strategies $ada2$
and $ada4$ try to keep the resource limit as low as possible, and each
evaluation is therefore very fast. However, once there are no improvements
possible with the current limit value, these strategies are slower to react.

The minimal limit value $minLimit$ and the maximum limit value $maxLimit$
are set to 500 and 15,000, respectively, based on the experience we gained from
our previous work~\cite{iccad17}.

\begin{table}[ht!]
\renewcommand{\arraystretch}{1.1}
\setlength{\tabcolsep}{5pt}
\centering
\caption{Adaptive strategy parameters.}\
\begin{tabular}{|c|cccccc|}\hline
Strategy      & \textbf{$\tau_{dec}$} & \textbf{$\tau_{inc}$} & \textbf{$\tau_{res}$} & period & minLimit & maxLimit \\ \hline
\textbf{ada1} & 4                   & 2                   & 10                  & 1000 & 500 & 15000  \\
\textbf{ada2} & 2                   & 1                   & 5                   & 15000 & 500 & 15000   \\
\textbf{ada3} & 4                   & 4                   & 8                   & 3000 & 500 & 15000   \\
\textbf{ada4} & 1                   & 1                   & 3                   & 5000 & 500 & 15000   \\
\textbf{ada5} & 5                   & 4                   & 8                   & 5000 & 500 & 15000  \\\hline
\end{tabular}
\label{table-ada-params}
\end{table}


%

\new{We evaluate the performance of the described strategies on the approximation scenario of 16-bit multipliers with a total of 8 target WCAE values ranging from $10^{-4} \%$ to $1 \%$ with 50 independent 1 hour and 6 hour long evolutionary runs. The quality of the obtained final solutions was evaluated using Friedman and Nemenyi statistical tests with results illustrated in Table~\ref{table:adaptive5}.

For 1h runs, stategy $ada5$ performs the best, but it's performance is statistically equivalent to $ada4$ and $ada3$. This group of strategies is significantly better than $ada1$ and $ada2$.

For 6h runs, $ada2$ significantly outperforms the rest of strategies, followed by $ada4$ which also significantly outperforms its successors.

In the overall evaluation, the group $ada2$, $ada4$ and $ada5$ is statistically equivalent and significantly better than strategies $ada1$ and $ada3$. As we aim to acquire the best solutions that our method can
provide, we select the strategies $ada2$ and $ada4$ as representatives of
adaptive strategies for the following experiments. }

\begin{table}[]
\renewcommand{\arraystretch}{1.2}
\setlength{\tabcolsep}{3pt}
\caption{\new{Pairwise p-values of Nemenyi statistical test and average rank values for 1h experiments, 6h experiments and combined.}}
\centering
\label{table:adaptive5}
\scalebox{0.8}{
\begin{tabular}{|c|lllll|}\hline
\multicolumn{1}{|l|}{1h runs} & \multicolumn{1}{c}{$ada1$} & \multicolumn{1}{c}{$ada2$} & \multicolumn{1}{c}{$ada3$} & \multicolumn{1}{c}{$ada4$} & \multicolumn{1}{c|}{$ada5$} \\ \hline
$ada2$ & 0.74708 & - & - & - & - \\
$ada3$ & \textbf{6.50E-06} & \textbf{0.00155} & - & - & - \\
$ada4$ & \textbf{4.30E-07} & \textbf{0.00019} & 0.98709 & - & - \\
$ada5$ & \textbf{9.20E-13} & \textbf{4.10E-09} & 0.09467 & 0.27627 & - \\
Rank & 3.4275 & 3.2925 & 2.87125 & 2.815 & 2.59375 \\ \hline\hline
\multicolumn{1}{|l|}{6h runs} & \multicolumn{1}{c}{$ada1$} & \multicolumn{1}{c}{$ada2$} & \multicolumn{1}{c}{$ada3$} & \multicolumn{1}{c}{$ada4$} & \multicolumn{1}{c|}{$ada5$} \\ \hline
$ada2$ & \textbf{2.00E-16} & - & - & - & - \\
$ada3$ & 0.28188 & \textbf{4.90E-14} & - & - & - \\
$ada4$ & \textbf{5.40E-14} & \textbf{0.01082} & \textbf{3.10E-12} & - & - \\
$ada5$ & \textbf{0.00013} & \textbf{6.50E-14} & 0.12034 & \textbf{9.10E-06} & - \\
Rank & 3.6275 & 2.23125 & 3.4075 & 2.5925 & 3.14125 \\ \hline\hline
\multicolumn{1}{|l|}{Overall} & \multicolumn{1}{c}{$ada1$} & \multicolumn{1}{c}{$ada2$} & \multicolumn{1}{c}{$ada3$} & \multicolumn{1}{c}{$ada4$} & \multicolumn{1}{c|}{$ada5$} \\ \hline
$ada2$ & \textbf{6.10E-14} & - & - & - & - \\
$ada3$ & \textbf{9.00E-06} & \textbf{1.80E-05} & - & - & - \\
$ada4$ & \textbf{5.20E-14} & 0.9483 & \textbf{3.60E-07} & - & - \\
$ada5$ & \textbf{5.30E-14} & 0.6686 & \textbf{0.0053} & 0.2326 & - \\
Rank & 3.5275 & 2.761875 & 3.139375 & 2.70375 & 2.8675\\\hline
\end{tabular}}
\end{table}

\begin{figure}

\includegraphics[width=\columnwidth]{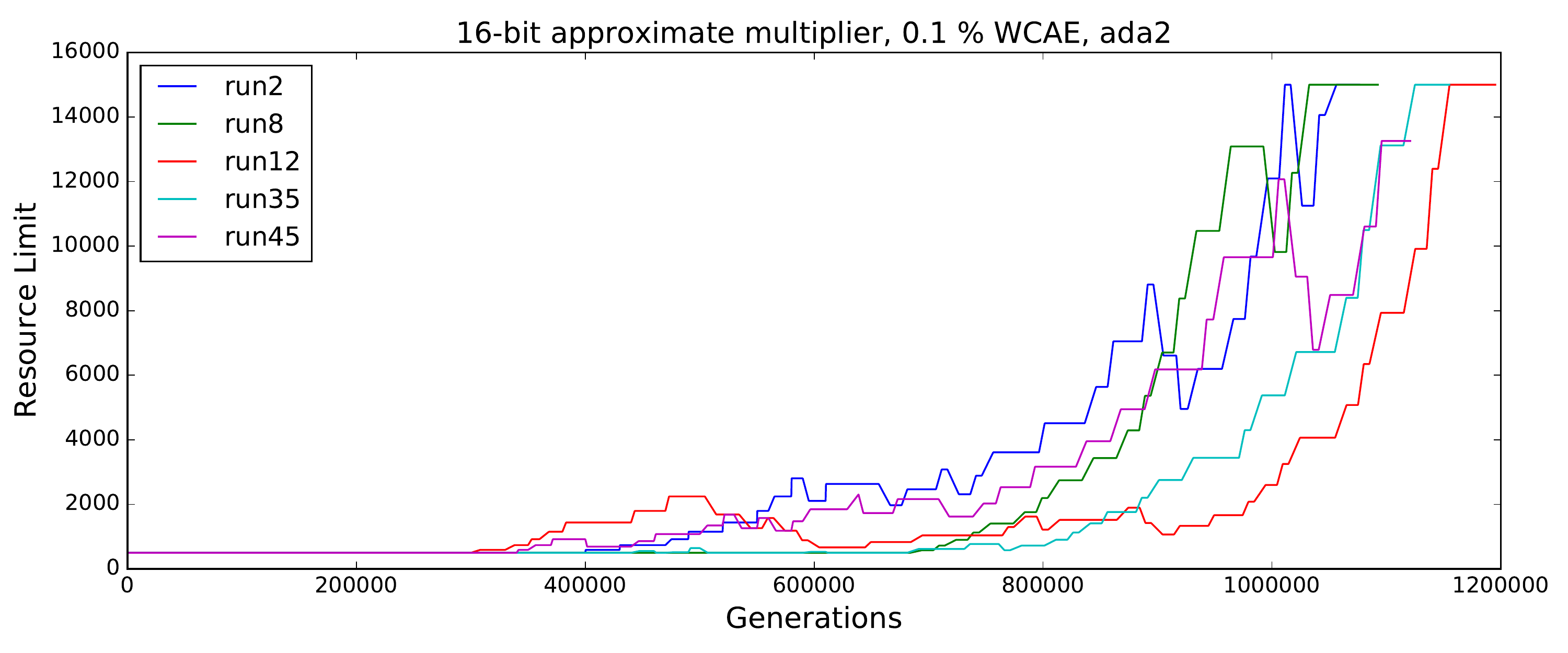}
\includegraphics[width=\columnwidth]{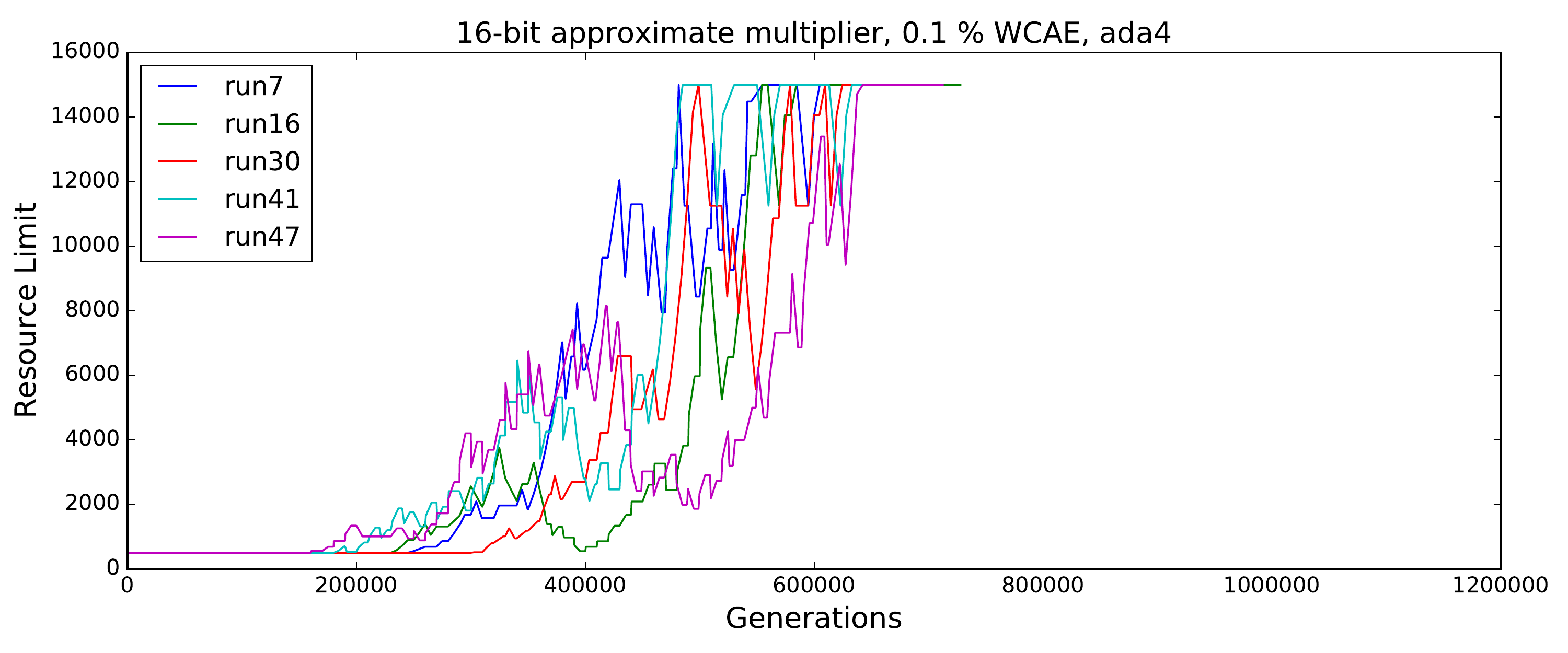}
\includegraphics[width=\columnwidth]{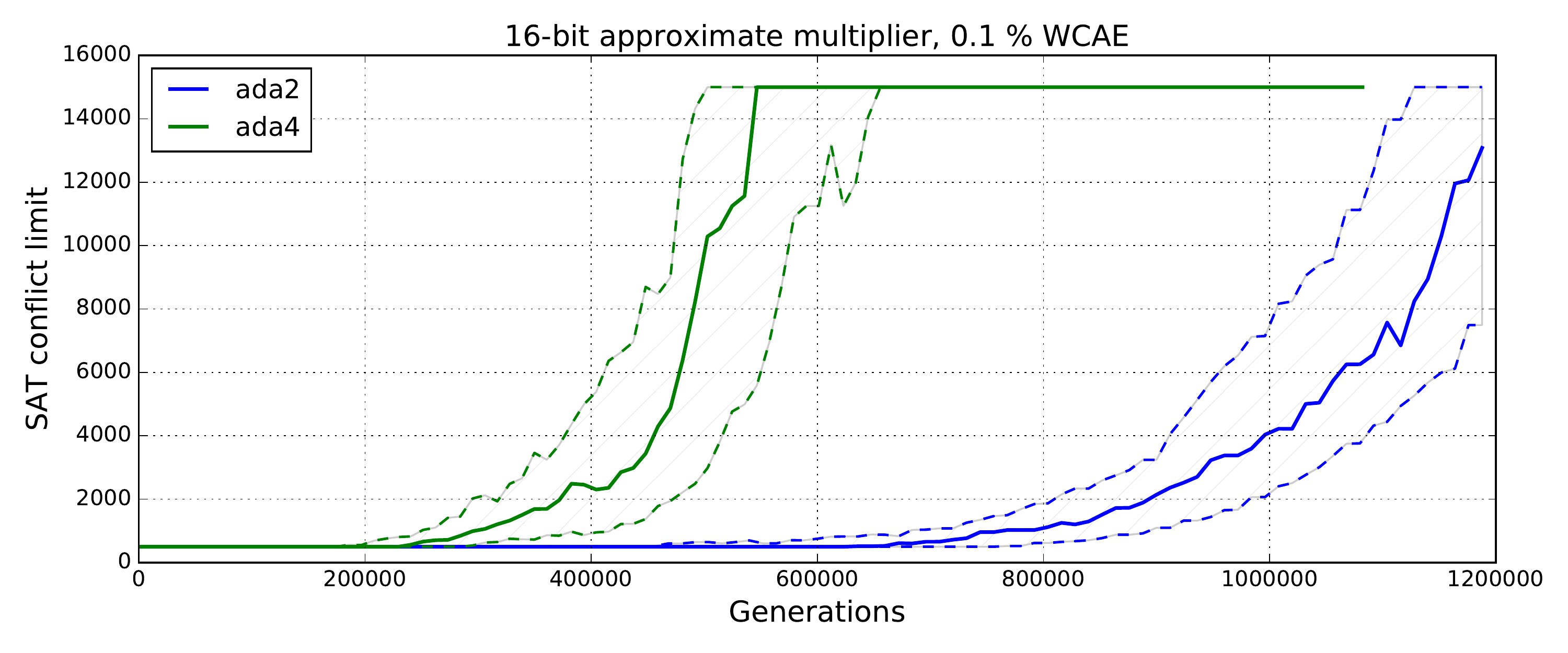}
\caption{The resource limits chosen by the adaptive strategies $ada2$ and $ada4$ during the approximation of the 16-bit multiplier. The top two plots illustrate five selected runs. The bottom plot shows the medians (full lines) and the quartiles Q1 and Q3 (dotted lines) over 50 runs.}
\label{lim-progress-mul}
\end{figure}

We further show how the adaptive strategies $ada2$ and $ada4$ change the
resource limits during the approximation process. Fig.~\ref{lim-progress-mul}
shows how the limits change (increase as well as decrease) over the time during
the approximation of the 16-bit multipliers with target $0.1\,\%$ WCAE. The
approximation ran for 6 hours, and the plot shows the maximum number of SAT
backtracks (i.e. the resource limit) that was allowed to be used during the
verification of candidate circuits in particular generations of the evolution
optimisation. The top two plots of the figure illustrate five selected runs for
both strategies, and the bottom plot shows the aggregated results for 50
independent runs. It shows the median of the resource limits plotted by the full
lines and quartiles Q1 and Q3 plotted by the dashed lines.


The figure confirms our expectations: in the initial stages of the
approximation, the limit is kept low because improvements are found frequently.
We can also see that the limit increases as well as decreases, and a closer
evaluation of our data reveals that both the periodic and immediate decrease are
used. Further, note that $ada4$ increases the limit much sooner than $ada2$,
and the rate of the increase is also much steeper. This fact allows $ada4$
to use more time out of the total time available for the entire evolutionary run
for evolving and evaluating solutions that need larger resource limits for their
verification. On the other hand, the higher limit slows the evolutionary
process down significantly---we see that none of the $ada4$ runs reaches the
number of $1.2 * 10^6$ generations in this experiment. The particular runs of
the strategies also demonstrate that $ada4$ exhibits more changes (including
periodic drops of the limits) compared with the more stable strategy $ada2$.
The impact of these differences on the quality of the obtained final
solutions will be evaluated in the further subsections of this section.

\begin{figure}[h]

\includegraphics[width=\columnwidth]{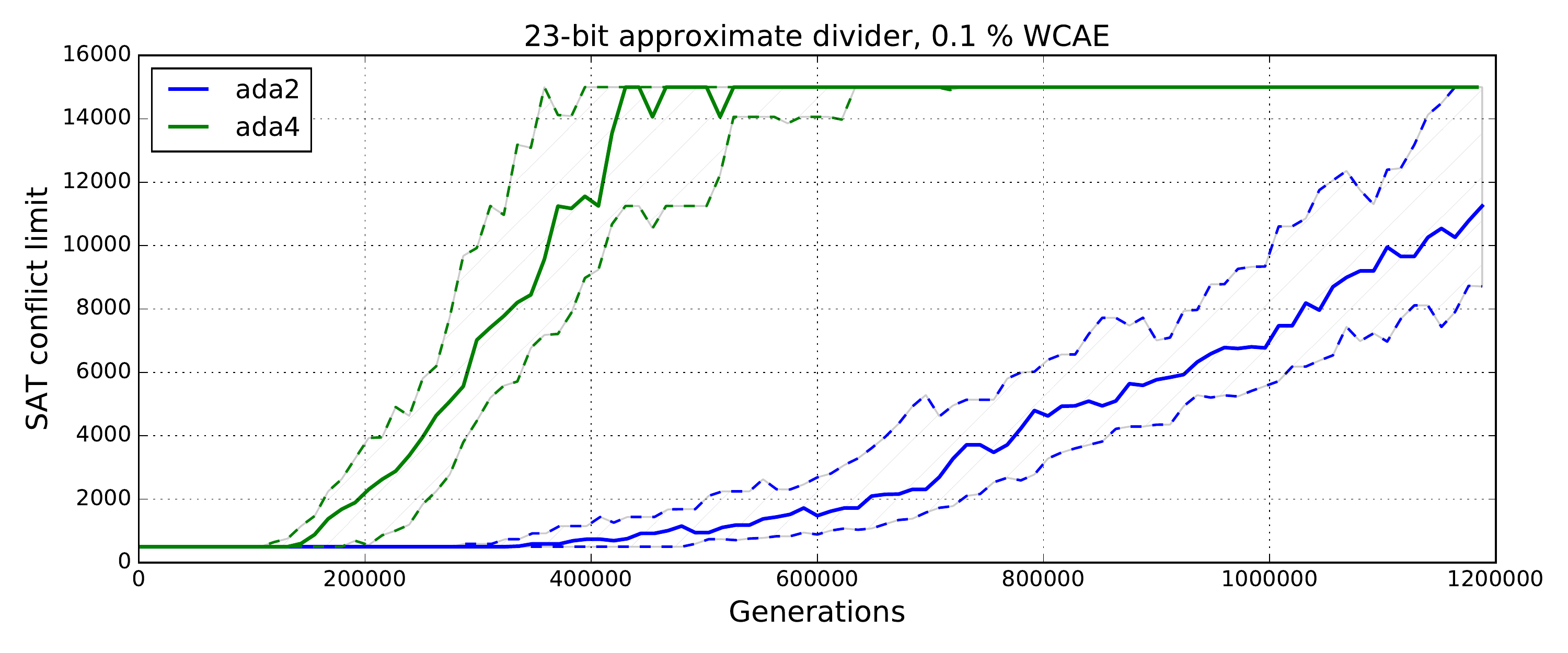}
\caption{The resource limits chosen by the adaptive strategies $ada2$ and $ada4$
during the approximation of a 23-bit divider. The plot shows the median values
(full lines) and the quartiles Q1 and Q3 (dotted lines) over 50 runs.}
\label{lim-progress-div}
\end{figure}

Finally, Fig.~\ref{lim-progress-div} shows the aggregated results for
approximation of 23-bit dividers (with the same WCAE) representing a very
different approximation scenario. We observe that the approximation of the
dividers requires higher resource limits (i.e. more time for the verification of
the candidate solutions) when compared with the multipliers: this is due to the
structural complexity of the circuits. For example, in the 400K-th generation,
$ada2$ sets the limit to about 2K for the multipliers and to about 12K for the
dividers. Note that the difference is, however, less significant in the case
of $ada4$.

\subsection{Reduction of randomness (Q1)}

Evolutionary algorithms involve a significant amount of randomness, and the
quality of the final solutions produced by independent runs can considerably vary.
One of the goals of the newly designed adaptive strategies is to reduce the
amount of the involved randomness and ensure that most of the approximation
runs will lead to high-quality solutions. In this part of experiments, we
examine the quality and variability of sets of 50 independent evolutionary runs
for the adaptive strategies as well as for various fix-limit resource settings. 

In the following experiments, we denote the fixed resource limit strategies as
$lim100$, $lim2K$, $lim10K$, $lim20K$, and $lim50K$ for the resource limits of
$100$, $2000$, $10000$, $20000$, and $50000$ backtracks on a single variable,
which are used through the whole evolutionary process. We chose these values to
represent small, mid range, and large values. In our previous
work~\cite{iccad17}, we used $lim20K$ as the standard resource limit~setting. 

\begin{figure}
\centering
\includegraphics[width=0.45\columnwidth]{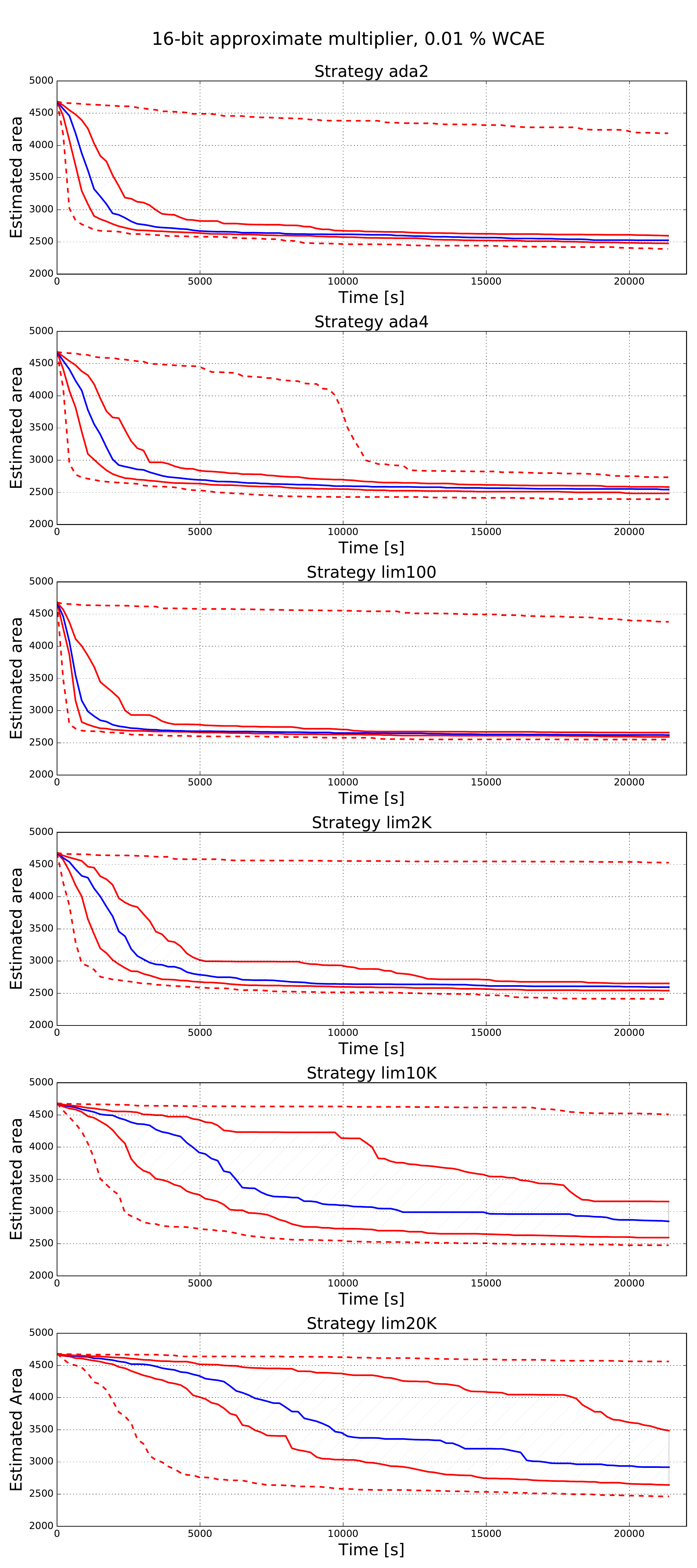}\\
\includegraphics[width=0.45\columnwidth]{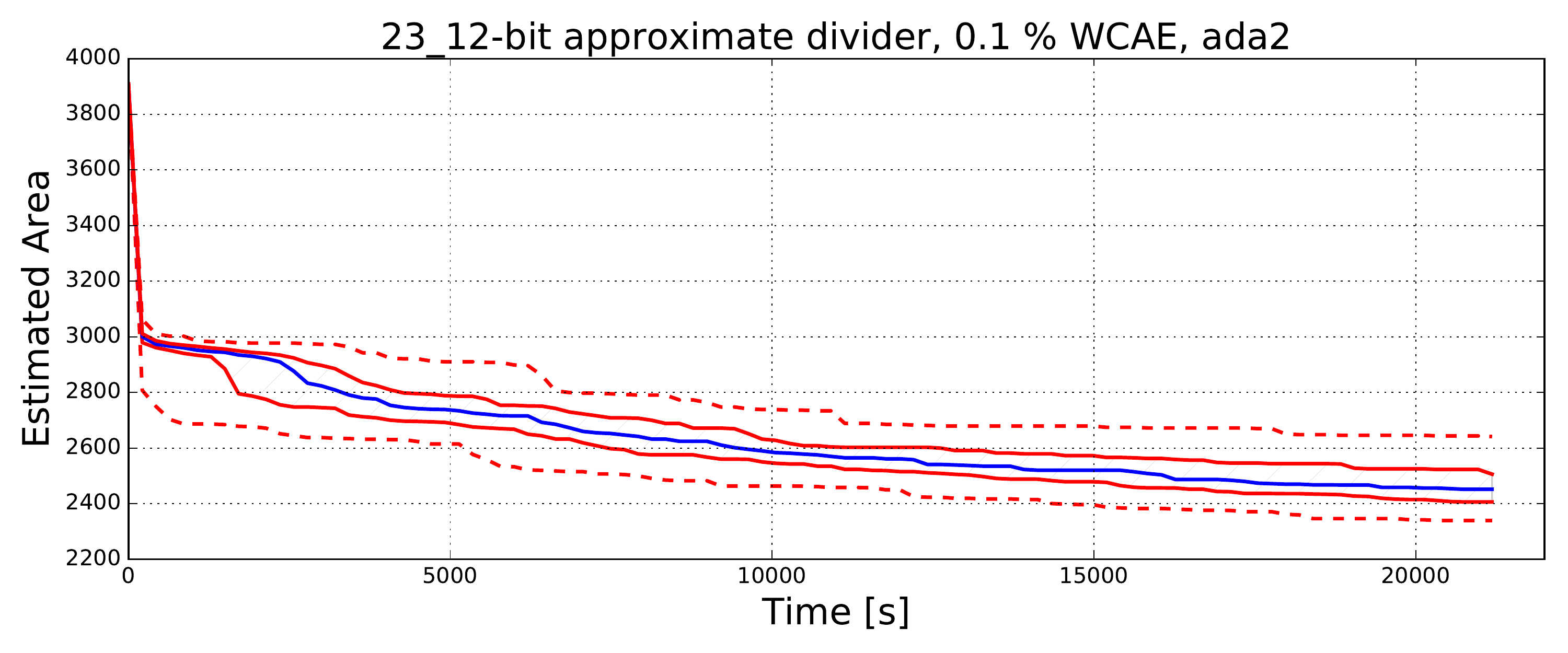}
\vspace{-1.5em}
\caption{Convergence curves for resource limit strategies showing the estimated
area for the best, worst, Q1, Q3, and median solutions during 16-bit multiplier and 23-bit divider
approximation.}
\label{ada-lim-robustness}
\end{figure}

The plots in Fig.~\ref{ada-lim-robustness} demonstrate how the size of the
candidate solutions decreases during particular runs. In particular, the dashed
red lines show the best and the worst run; and median, first (Q1) and third (Q3)
quartile are illustrated by the full blue line and the red lines, respectively.

The figure shows that the adaptive strategies as well as the strategies with
lower resource limit values are significantly more stable than the strategies
with higher limits. This is caused by the fact that the evolution has to explore
solutions requiring a long verification time. Such solutions are immediately
refused by the lower resource limits ($100$, $2K$) and by the adaptive
strategies but more likely accepted by the other strategies ($10K$, $20K$, and
$50K$). The long evaluation time is inherited from parents to offsprings. The
strategies with higher limit settings are therefore much slower to converge to a
near optimum solution. The previously described slowdown of the evolution also
leads to higher variation in the candidate quality throughout the evolution,
which can be observed as the wide interquartile range (IQR) for limits $10K$,
$20K$, and $50K$. Other strategies feature a narrow IQR---a desirable attribute
of a good resource limit strategy. The convergence of the strategy $lim50K$ is
so slow that we exclude this strategy from the rest of the experiments to save
computational time.

We obtained similar observations for other WCAE values and bit-width settings
for \new{multipliers, MACs and square circuits}. The difference between strategies is even more
pronounced for smaller approximation errors, which represent a harder
optimization problem. On the other hand, large approximation errors diminish the
differences.

The approximation of dividers represents another class of optimization problems
with a different behaviour. The variance of the solutions is very similar for all resource limit settings. This fact
is illustrated in the bottom part of Fig.~\ref{ada-lim-robustness}. While the variance is almost
identical, what differs between the resource limit strategies is the quality of
final solutions that can be achieved. This is described in greater detail in the
next section.

\vspace{1em}
\noindent
\new{\textbf{Summary for Q1:} For a wide class of circuits, the adaptive strategies as well as the low-limit strategies are significantly more stable than other fixed limit strategies (i.e. the effect of the randomness is smaller). All strategies show good stability for the approximation of dividers, however, the low-limit strategies provide considerably smaller reductions of the circuit area.}

\vspace{.5em}
\noindent
\emph{Note:} Since it would be very difficult to present our results while also showing the
randomness of the evolutionary runs at the same time, we present only the
quality of the median solutions in the rest of our paper when not stated
otherwise.
\subsection{Versatility of adaptive strategies (Q2)}

\begin{figure}
\centering
\includegraphics[width=0.7\columnwidth]{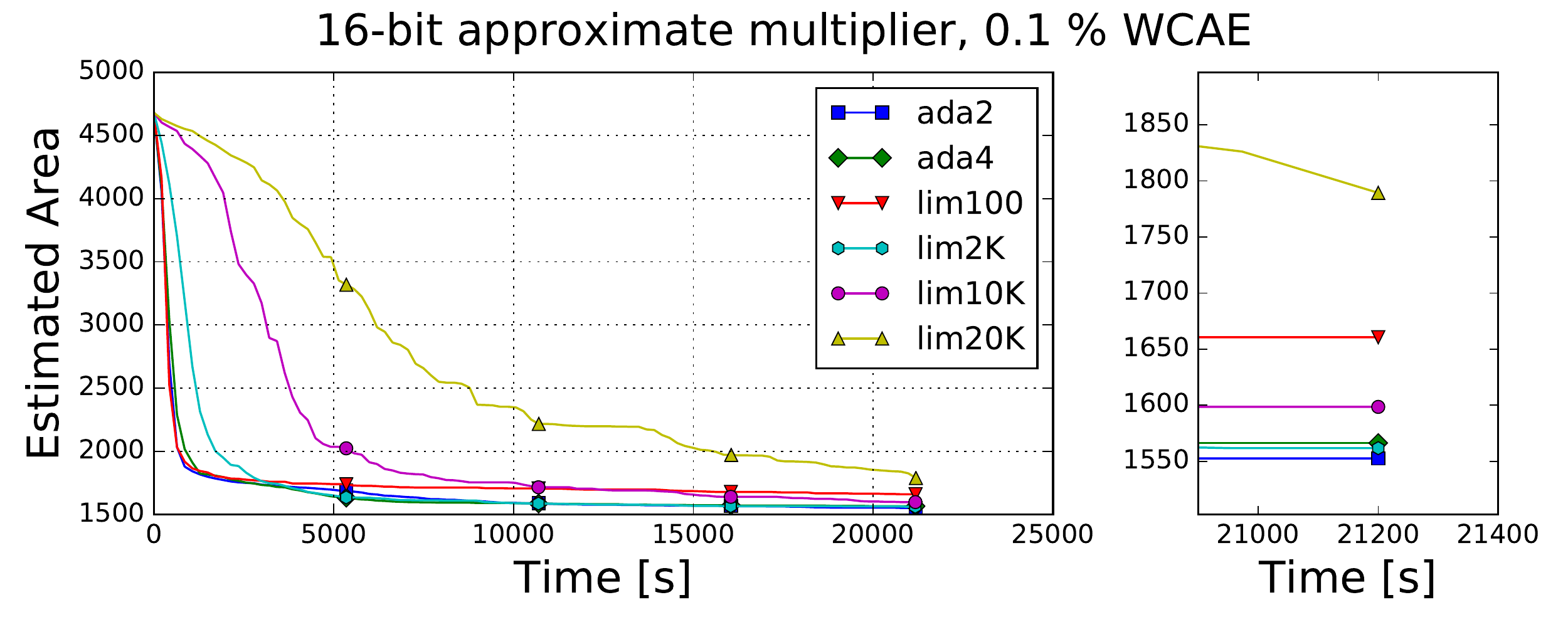}
\includegraphics[width=0.7\columnwidth]{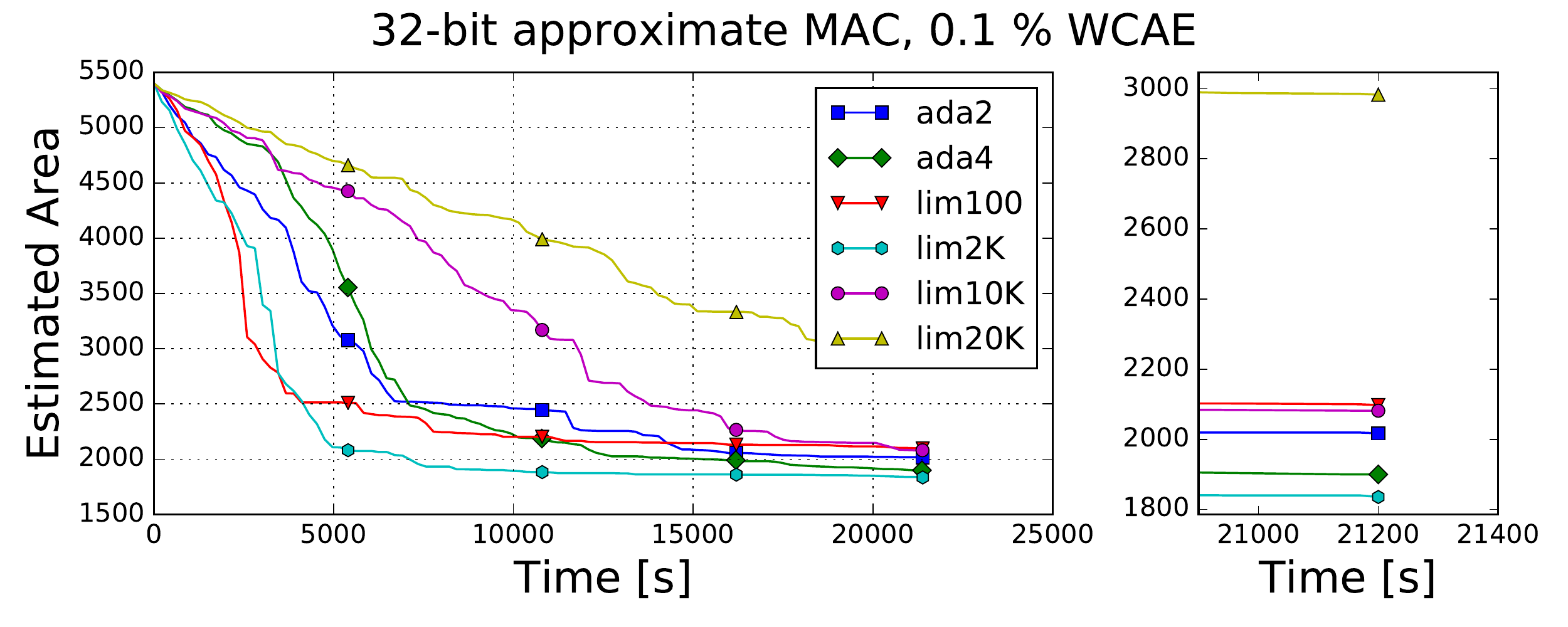}
\includegraphics[width=0.7\columnwidth]{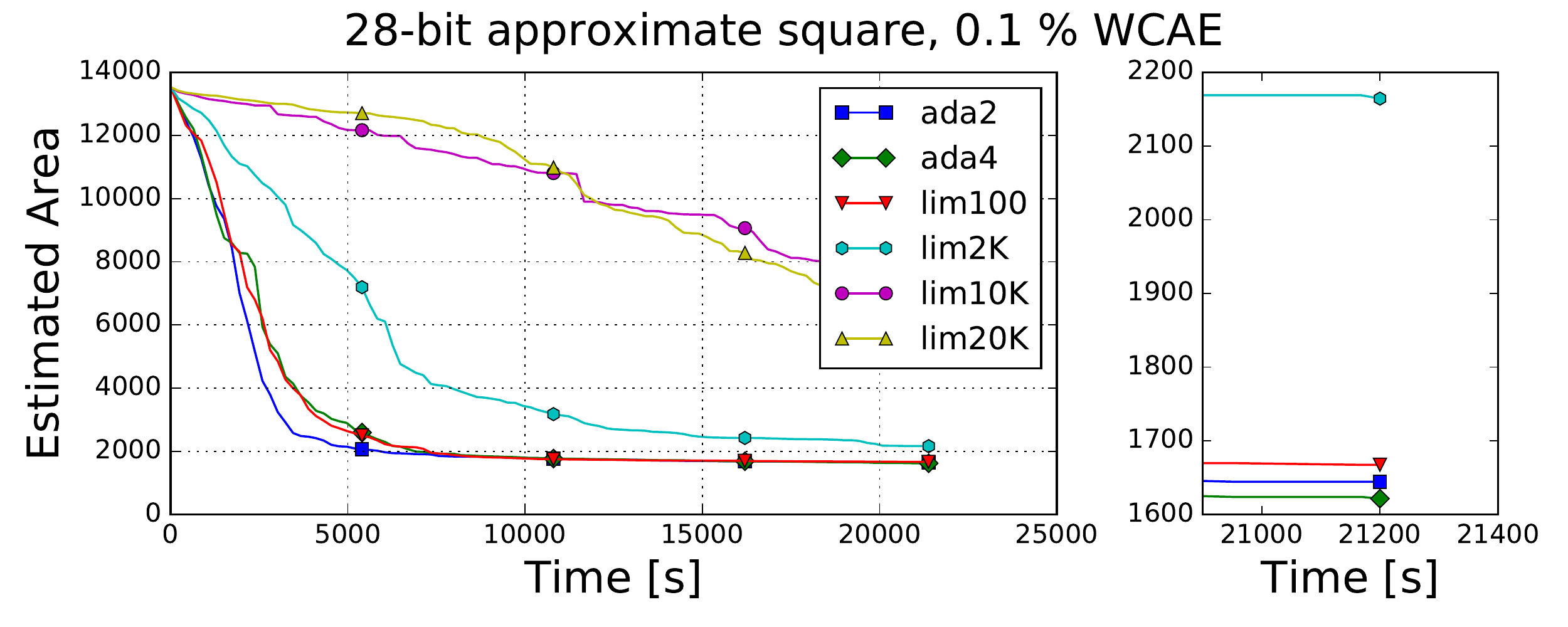}
\includegraphics[width=0.7\columnwidth]{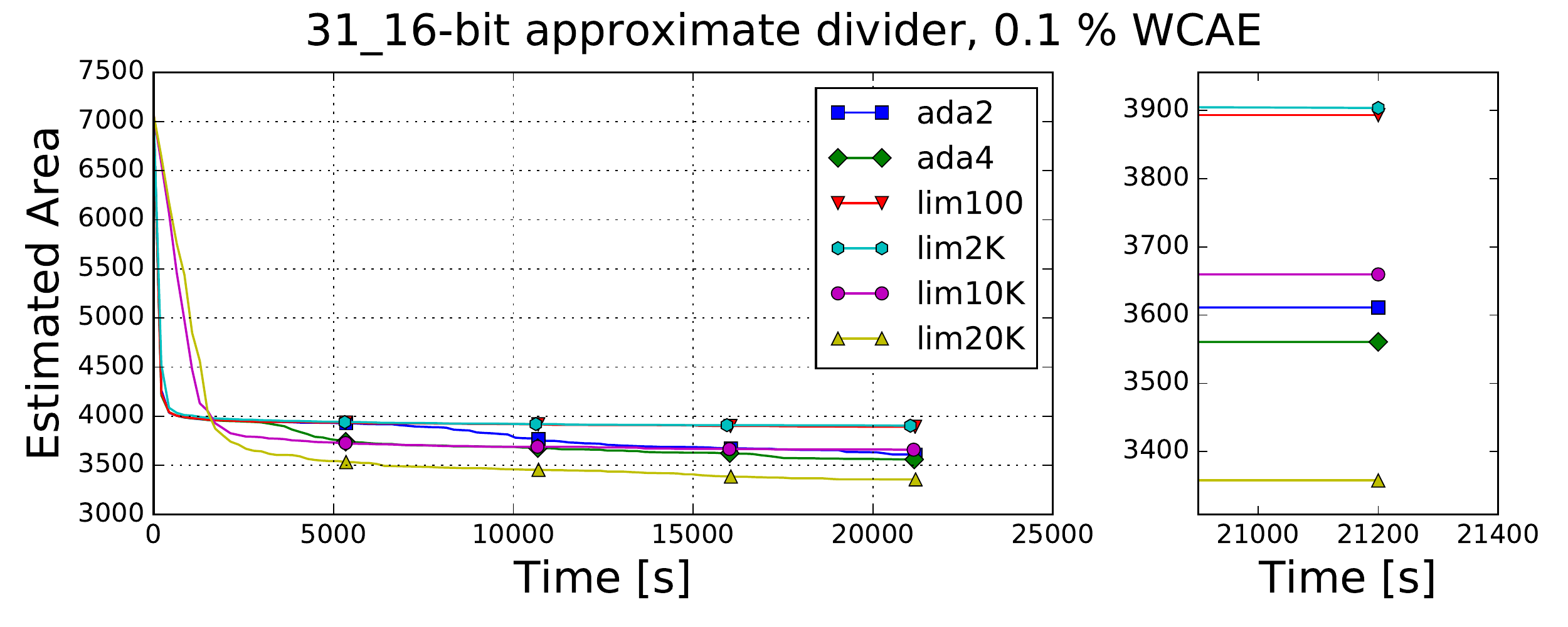}
\caption{Convergence plots of median solutions calculated from 50 independent
evolutionary runs for various combinational circuits.}
\label{ada-vs-lim}
\vspace{-1em}
\end{figure}

The key feature of circuit approximation strategies is \emph{versatility}, an
ability to provide excellent performance for various approximation scenarios
including different circuits, WCAE values, and time limits.  Although the
verifiability-driven strategy itself leads to unprecedented performance and
scalability of circuit approximation~\cite{iccad17}, the fixed-limit resource
limits do not ensure versatility. This fact is demonstrated in
Fig.~\ref{ada-vs-lim} where we fix WCAE to $0.1~\%$ for \new{multipliers, squares, MACs, and
dividers}, and explore the progress of the approximation process. \new{The right part of each plot illustrates the quality of the final solutions.}

When comparing the performance of the fixed-limit strategies on the
approximation of 16-bit multipliers, we can see that the strategy $lim100$
dominates in the first hour of the approximation process since it provides the
fastest convergence. Strategies $ada2$, $ada4$, and $lim2K$ converge slower, but
around after the first hour their median solutions outperform $lim100$ which
cannot achieve further improvements due to the tight resource limit. Strategies
$lim10K$ and $lim20K$ provide a significantly slower converge: $lim10K$ needs
around 2.5 hours to provide solutions that are comparable to the aforementioned
strategies, $lim20K$ is too slow and its final solution  lags
behind.

Similar trends among the inspected strategies are observed for the 32-bit MACs
(see the \new{second} plot in Fig.~\ref{ada-vs-lim}). Note that, in general, the
convergence is much slower because this circuit is larger and represents a
harder optimization problem compared with the multipliers. Moreover, we observe
a larger diversity among the strategies.

\new{The progress tendencies for \mbox{28-bit} square circuit (see the third plot in Fig.~\ref{ada-vs-lim}) significantly differ. The strategies
$lim10K$ and $lim20K$ provide an extremely slow converge and even after 6-hours runs they significantly lag behind the other strategies.
The strategies $lim2K$ also converges much slower than the remaining strategies, which show similar performance. After an 1-hour run, $lim2K$ returns circuits that are about two-times larger than the circuits provided by the strategy $lim100$, however,  after 5 hours it catches up with the other strategies.}

The bottom part of Fig.~\ref{ada-vs-lim} illustrates results for the dividers.
We observe a very different trend in the approximation process. In particular,
all strategies converge very quickly to a~sub-optimal solution, but the
fixed-limit strategies with small resource limits ($lim100$ and $lim2K$) are not
able to achieve any further improvement, and they significantly lag behind other
strategies in the final solutions. We further observe that the strategy
$lim20K$, which performes very poorly on the previous circuits, is the best
strategy in this case. The proposed adaptive strategies inherit the initial fast
convergence using a~small limit, but they adapt the limit after the first hour
and arrive to results comparable with the strategy $lim10K$.

Fig.~\ref{ada-vs-lim} indicates that the performance of the particular
fixed-limit strategies fundamentally varies for different circuits under
approximation. For example, the strategy $lim2K$ gives the best results for the
MACs, but it behaves very poorly on the dividers, which clearly require a very
high resource limit. In Tables~\ref{table:mult}--\ref{table:square}, depicting the results for particular circuits, we show that
the selection of the best strategy also depends on the required  WCAE and on the
bit-width of the particular circuits. The tables list the relative size
reductions of the median solutions with respect to the golden circuit obtained
using different strategies after 1 and 6 hours  for different circuit types,
bit-widths, WCAEs. The best solution for each target approximation error is
highlighted in bold text. For instance, Table~\ref{table:mult} shows that the
median solution for 16-bit multipliers with $0.01~\%$ WCAE obtained by $ada4$ in
6 hours has the area of $54~\%$ of the original 16-bit multiplier. The quality
of this solution dominates the solutions obtained by other strategies for this
experimental setup.

\begin{table}[]
\renewcommand{\arraystretch}{1.1}
\setlength{\tabcolsep}{4pt}
\caption{The relative sizes in \% of median solutions with respect to the size of the golden solution for multiplier approximation.}
\label{table:mult}
\centering

\resizebox{0.45\linewidth}{!}{
\begin{tabular}{rrrrrrr}
\multicolumn{2}{l}{16-bit multiplier} & \multicolumn{1}{l}{} & \multicolumn{1}{l}{} & \multicolumn{1}{l}{} & \multicolumn{1}{l}{} & \multicolumn{1}{l}{} \\ \hline
\multicolumn{1}{|c|}{1h runs} & \multicolumn{1}{c}{$lim100$} & \multicolumn{1}{c}{$lim2K$} & \multicolumn{1}{c}{$lim10K$} & \multicolumn{1}{c}{$lim20K$} & \multicolumn{1}{c}{$ada2$} & \multicolumn{1}{c|}{$ada4$} \\ \hline
\multicolumn{1}{|r|}{0.001 \%} & 82.2 & 85.8 & 95.7 & 98.1 & 82.4 & \multicolumn{1}{r|}{\textbf{81.1}} \\
\multicolumn{1}{|r|}{0.01 \%} & 61.1 & 60.4 & 88.6 & 96.7 & 59.3 & \multicolumn{1}{r|}{\textbf{57.9}} \\
\multicolumn{1}{|r|}{0.1 \%} & 37.7 & 37.0 & 58.9 & 86.2 & 36.8 & \multicolumn{1}{r|}{\textbf{36.7}} \\
\multicolumn{1}{|r|}{1 \%} & 18.8 & 17.9 & 20.2 & 43.5 & 18.6 & \multicolumn{1}{r|}{\textbf{17.7}} \\ \hline
\multicolumn{1}{|c|}{6h runs} & \multicolumn{1}{l}{$lim100$} & \multicolumn{1}{l}{$lim2K$} & \multicolumn{1}{l}{$lim10K$} & \multicolumn{1}{l}{$lim20K$} & \multicolumn{1}{l}{$ada2$} & \multicolumn{1}{l|}{$ada4$} \\ \hline
\multicolumn{1}{|r|}{0.001 \%} & 74.0 & 72.8 & 77.6 & 82.4 & 72.5 & \multicolumn{1}{r|}{\textbf{71.5}} \\
\multicolumn{1}{|r|}{0.01 \%} & 56.4 & 55.4 & 56.8 & 63.2 & 55.0 & \multicolumn{1}{r|}{\textbf{54.0}} \\
\multicolumn{1}{|r|}{0.1 \%} & 35.5 & 33.4 & 34.6 & 38.5 & \textbf{33.2} & \multicolumn{1}{r|}{33.5} \\
\multicolumn{1}{|r|}{1 \%} & 17.4 & \textbf{15.7} & 16.4 & 17.2 & \textbf{15.7} & \multicolumn{1}{r|}{15.9} \\ \hline
\vspace{-0.5em}
\end{tabular}
}
\quad
\resizebox{0.45\linewidth}{!}{
\begin{tabular}{rrrrrrr}
\multicolumn{2}{l}{24-bit multiplier} & \multicolumn{1}{l}{} & \multicolumn{1}{l}{} & \multicolumn{1}{l}{} & \multicolumn{1}{l}{} & \multicolumn{1}{l}{} \\ \hline
\multicolumn{1}{|c|}{1h runs} & \multicolumn{1}{c}{$lim100$} & \multicolumn{1}{c}{$lim2K$} & \multicolumn{1}{c}{$lim10K$} & \multicolumn{1}{c}{$lim20K$} & \multicolumn{1}{c}{$ada2$} & \multicolumn{1}{c|}{$ada4$} \\ \hline
\multicolumn{1}{|r|}{0.001 \%} & 91.2 & 87.6 & 96.7 & 97.7 & 89.0 & \multicolumn{1}{r|}{\textbf{87.2}} \\
\multicolumn{1}{|r|}{0.01 \%} & 32.1 & 59.7 & 89.3 & 94.4 & 32.8 & \multicolumn{1}{r|}{\textbf{31.5}} \\
\multicolumn{1}{|r|}{0.1 \%} & 19.0 & 19.8 & 78.7 & 86.3 & \textbf{18.4} & \multicolumn{1}{r|}{18.5} \\
\multicolumn{1}{|r|}{1 \%} & 9.2 & \textbf{8.6} & 21.1 & 79.9 & 9.1 & \multicolumn{1}{r|}{8.9} \\ \hline
\multicolumn{1}{|c|}{6h runs} & \multicolumn{1}{c}{$lim100$} & \multicolumn{1}{c}{$lim2K$} & \multicolumn{1}{c}{$lim10K$} & \multicolumn{1}{c}{$lim20K$} & \multicolumn{1}{c}{$ada2$} & \multicolumn{1}{c|}{$ada4$} \\ \hline
\multicolumn{1}{|r|}{0.001 \%} & 43.0 & \textbf{40.4} & 79.3 & 82.6 & 41.8 & \multicolumn{1}{r|}{41.0} \\
\multicolumn{1}{|r|}{0.01 \%} & 27.1 & 27.1 & 30.2 & 42.7 & 26.8 & \multicolumn{1}{r|}{\textbf{26.3}} \\
\multicolumn{1}{|r|}{0.1 \%} & 16.3 & \textbf{15.9} & 18.1 & 23.4 & \textbf{15.9} & \multicolumn{1}{r|}{16.0} \\
\multicolumn{1}{|r|}{1 \%} & 8.7 & 7.6 & 7.6 & 8.6 & \textbf{7.4} & \multicolumn{1}{r|}{\textbf{7.4}} \\ \hline
\end{tabular}
}
\end{table}

\begin{table}[]
\setlength{\tabcolsep}{5pt}
\renewcommand{\arraystretch}{1.08}
\setlength{\tabcolsep}{4pt}
\caption{The relative sizes in \% of median solutions with respect to the size
of the golden solution for MAC approximation and divider approximation.}
\label{table:mac}
\centering
\resizebox{0.45\linewidth}{!}{
\begin{tabular}{rrrrrrr}
\multicolumn{2}{l}{24-bit MAC} & \multicolumn{1}{l}{} & \multicolumn{1}{l}{} & \multicolumn{1}{l}{} & \multicolumn{1}{l}{} & \multicolumn{1}{l}{} \\ \hline
\multicolumn{1}{|c|}{1h runs} & \multicolumn{1}{c}{$lim100$} & \multicolumn{1}{c}{$lim2K$} & \multicolumn{1}{c}{$lim10K$} & \multicolumn{1}{c}{$lim20K$} & \multicolumn{1}{c}{$ada2$} & \multicolumn{1}{c|}{$ada4$} \\ \hline
\multicolumn{1}{|l|}{0.0001 \%} & \textbf{96.7} & 96.9 & 97.7 & 97.8 & \textbf{96.7} & \multicolumn{1}{r|}{97.1} \\
\multicolumn{1}{|r|}{0.001 \%} & 94.3 & \textbf{93.7} & 95.0 & 95.3 & 94.0 & \multicolumn{1}{r|}{93.9} \\
\multicolumn{1}{|r|}{0.01 \%} & 91.3 & \textbf{82.1} & 83.0 & 95.5 & 90.0 & \multicolumn{1}{r|}{93.9} \\
\multicolumn{1}{|r|}{0.1 \%} & 73.1 & 67.8 & 93.2 & 92.6 & 65.8 & \multicolumn{1}{r|}{\textbf{64.1}} \\
\multicolumn{1}{|r|}{1 \%} & 38.2 & 28.9 & 45.4 & 67.9 & 31.1 & \multicolumn{1}{r|}{\textbf{28.5}} \\ \hline
\multicolumn{1}{|c|}{6h runs} & \multicolumn{1}{c}{$lim100$} & \multicolumn{1}{c}{$lim2K$} & \multicolumn{1}{c}{$lim10K$} & \multicolumn{1}{c}{$lim20K$} & \multicolumn{1}{c}{$ada2$} & \multicolumn{1}{c|}{$ada4$} \\ \hline
\multicolumn{1}{|l|}{0.0001 \%} & 95.9 & 96.2 & 95.8 & 96.2 & \textbf{95.5} & \multicolumn{1}{r|}{95.8} \\
\multicolumn{1}{|r|}{0.001 \%} & 92.3 & 90.4 & 89.4 & 89.4 & 88.7 & \multicolumn{1}{r|}{\textbf{88.5}} \\
\multicolumn{1}{|r|}{0.01 \%} & 84.9 & 76.0 & 75.0 & 78.7 & 76.6 & \multicolumn{1}{r|}{\textbf{75.2}} \\
\multicolumn{1}{|r|}{0.1 \%} & 59.1 & 56.7 & 61.2 & 65.6 & 53.1 & \multicolumn{1}{r|}{\textbf{53.0}} \\
\multicolumn{1}{|r|}{1 \%} & 31.8 & 27.3 & 26.1 & 26.9 & \textbf{24.7} & \multicolumn{1}{r|}{24.9} \\ \hline
\end{tabular}
}
\quad
\resizebox{0.45\linewidth}{!}{
\begin{tabular}{rrrrrrr}
\multicolumn{2}{l}{32-bit MAC} & \multicolumn{1}{l}{} & \multicolumn{1}{l}{} & \multicolumn{1}{l}{} & \multicolumn{1}{l}{} & \multicolumn{1}{l}{} \\ \hline
\multicolumn{1}{|c|}{1h runs} & \multicolumn{1}{c}{$lim100$} & \multicolumn{1}{c}{$lim2K$} & \multicolumn{1}{c}{$lim10K$} & \multicolumn{1}{c}{$lim20K$} & \multicolumn{1}{c}{$ada2$} & \multicolumn{1}{c|}{$ada4$} \\ \hline
\multicolumn{1}{|l|}{0.0001 \%} & \textbf{94.6} & 95.7 & 98.7 & 98.9 & 95.0 & \multicolumn{1}{r|}{94.8} \\
\multicolumn{1}{|r|}{0.001 \%} & 94.3 & 94.0 & 97.5 & 97.6 & 93.8 & \multicolumn{1}{r|}{\textbf{93.6}} \\
\multicolumn{1}{|r|}{0.01 \%} & 87.1 & \textbf{81.2} & 90.0 & 95.5 & 85.5 & \multicolumn{1}{r|}{89.3} \\
\multicolumn{1}{|r|}{0.1 \%} & 87.1 & \textbf{57.8} & 85.7 & 90.8 & 77.3 & \multicolumn{1}{r|}{86.2} \\
\multicolumn{1}{|r|}{1 \%} & 24.8 & \textbf{19.1} & 32.0 & 58.3 & 19.4 & \multicolumn{1}{r|}{19.7} \\ \hline
\multicolumn{1}{|c|}{6h runs} & \multicolumn{1}{c}{$lim100$} & \multicolumn{1}{c}{$lim2K$} & \multicolumn{1}{c}{$lim10K$} & \multicolumn{1}{c}{$lim20K$} & \multicolumn{1}{c}{$ada2$} & \multicolumn{1}{c|}{$ada4$} \\ \hline
\multicolumn{1}{|l|}{0.0001 \%} & 93.7 & 88.9 & 93.7 & 94.3 & 91.2 & \multicolumn{1}{r|}{\textbf{88.0}} \\
\multicolumn{1}{|r|}{0.001 \%} & 91.9 & 78.0 & 83.4 & 80.8 & 80.5 & \multicolumn{1}{r|}{\textbf{76.9}} \\
\multicolumn{1}{|r|}{0.01 \%} & 61.1 & 60.4 & 55.6 & 62.5 & \textbf{57.9} & \multicolumn{1}{r|}{62.7} \\
\multicolumn{1}{|r|}{0.1 \%} & 39.7 & \textbf{34.0} & 39.1 & 54.8 & 37.4 & \multicolumn{1}{r|}{35.1} \\
\multicolumn{1}{|r|}{1 \%} & 20.3 & 17.1 & 16.4 & 16.4 & 15.5 & \multicolumn{1}{r|}{\textbf{15.3}} \\ \hline
\end{tabular}
}

\vspace{1em}

\resizebox{0.45\linewidth}{!}{
\begin{tabular}{rrrrrrr}
\multicolumn{2}{l}{23-bit divider} & \multicolumn{1}{l}{} & \multicolumn{1}{l}{} & \multicolumn{1}{l}{} & \multicolumn{1}{l}{} & \multicolumn{1}{l}{} \\ \hline
\multicolumn{1}{|c|}{1h runs} & $lim100$ & $lim2K$ & $lim10K$ & $lim20K$ & $ada2$ & \multicolumn{1}{r|}{$ada4$} \\ \hline
\multicolumn{1}{|r|}{0.05 \%} & 76.8 & 76.6 & 76.4 & \textbf{74.2} & 76.7 & \multicolumn{1}{r|}{76.6} \\
\multicolumn{1}{|r|}{0.1 \%} & 72.7 & 71.1 & 66.7 & \textbf{65.1} & 71.3 & \multicolumn{1}{r|}{68.4} \\
\multicolumn{1}{|r|}{0.5 \%} & 51.4 & 48.2 & \textbf{43.1} & 43.4 & 48.9 & \multicolumn{1}{r|}{46.0} \\
\multicolumn{1}{|r|}{1 \%} & 42.3 & 37.2 & \textbf{33.2} & 34.6 & 39.8 & \multicolumn{1}{r|}{35.7} \\ \hline
\multicolumn{1}{|c|}{6h runs} & $lim100$ & $lim2K$ & $lim10K$ & $lim20K$ & $ada2$ & \multicolumn{1}{r|}{$ada4$} \\ \hline
\multicolumn{1}{|r|}{0.05 \%} & 73.7 & 74.3 & 72.4 & \textbf{69.9} & 72.6 & \multicolumn{1}{r|}{72.5} \\
\multicolumn{1}{|r|}{0.1 \%} & 66.5 & 67.8 & 63.9 & \textbf{61.4} & 62.7 & \multicolumn{1}{r|}{62.8} \\
\multicolumn{1}{|r|}{0.5 \%} & 47.8 & 44.0 & \textbf{38.6} & 39.9 & 40.1 & \multicolumn{1}{r|}{39.8} \\
\multicolumn{1}{|r|}{1 \%} & 39.0 & 32.2 & \textbf{29.4} & 30.2 & 30.3 & \multicolumn{1}{r|}{31.0} \\ \hline
\vspace{-0.5em}
\end{tabular}
}
\quad
\resizebox{0.45\linewidth}{!}{
\begin{tabular}{rrrrrrr}
\multicolumn{2}{l}{31-bit divider} & \multicolumn{1}{l}{} & \multicolumn{1}{l}{} & \multicolumn{1}{l}{} & \multicolumn{1}{l}{} & \multicolumn{1}{l}{} \\ \hline
\multicolumn{1}{|c|}{1h runs} & $lim100$ & $lim2K$ & $lim10K$ & $lim20K$ & $ada2$ & \multicolumn{1}{r|}{$ada4$} \\ \hline
\multicolumn{1}{|r|}{0.05 \%} & 62.4 & 62.5 & 63.1 & \textbf{60.5} & 62.3 & \multicolumn{1}{r|}{62.3} \\
\multicolumn{1}{|r|}{0.1 \%} & 55.8 & 56.0 & 53.3 & \textbf{51.1} & 55.8 & \multicolumn{1}{r|}{55.2} \\
\multicolumn{1}{|r|}{0.5 \%} & 42.5 & 38.4 & \textbf{31.6} & 29.8 & 38.3 & \multicolumn{1}{r|}{36.8} \\
\multicolumn{1}{|r|}{1 \%} & 33.9 & 28.3 & \textbf{21.6} & 22.6 & 31.5 & \multicolumn{1}{r|}{29.7} \\ \hline
\multicolumn{1}{|c|}{6h runs} & $lim100$ & $lim2K$ & $lim10K$ & $lim20K$ & $ada2$ & \multicolumn{1}{r|}{$ada4$} \\ \hline
\multicolumn{1}{|r|}{0.05 \%} & 61.8 & 62.0 & 60.5 & \textbf{58.2} & 59.1 & \multicolumn{1}{r|}{59.0} \\
\multicolumn{1}{|r|}{0.1 \%} & 55.1 & 55.2 & 51.8 & \textbf{47.5} & 51.2 & \multicolumn{1}{r|}{50.4} \\
\multicolumn{1}{|r|}{0.5 \%} & 37.9 & 37.2 & \textbf{27.8} & 26.9 & 30.6 & \multicolumn{1}{r|}{28.2} \\
\multicolumn{1}{|r|}{1 \%} & 30.4 & 26.0 & 19.1 & \textbf{19.0} & 22.0 & \multicolumn{1}{r|}{20.3} \\ \hline
\end{tabular}
}

\end{table}

\begin{table*}[]
\caption{The relative sizes in \% of median solutions with respect to the size of the golden solution for square approximation.}
\label{table:square}
\setlength{\tabcolsep}{4pt}
\renewcommand{\arraystretch}{1.1}
\centering

\resizebox{0.45\linewidth}{!}{
\begin{tabular}{rrrrrrr}
\multicolumn{2}{l}{20-bit square} & \multicolumn{1}{l}{} & \multicolumn{1}{l}{} & \multicolumn{1}{l}{} & \multicolumn{1}{l}{} & \multicolumn{1}{l}{} \\ \hline
\multicolumn{1}{|c|}{1h runs} & $lim100$ & $lim2K$ & $lim10K$ & $lim20K$ & $ada2$ & \multicolumn{1}{r|}{$ada4$} \\ \hline
\multicolumn{1}{|r|}{0.0001 \%} & \textbf{92.7} & 97.6 & 98.8 & 99.3 & 94.6 & \multicolumn{1}{r|}{93.4} \\
\multicolumn{1}{|r|}{0.001 \%} & 81.8 & 93.4 & 98.5 & 98.9 & 85.7 & \multicolumn{1}{r|}{\textbf{80.6}} \\
\multicolumn{1}{|r|}{0.01 \%} & \textbf{40.2} & 82.6 & 95.7 & 97.8 & 71.4 & \multicolumn{1}{r|}{51.2} \\
\multicolumn{1}{|r|}{0.1 \%} & 29.4 & \textbf{25.4} & 90.3 & 89.2 & 25.7 & \multicolumn{1}{r|}{26.9} \\
\multicolumn{1}{|r|}{1 \%} & 13.4 & 10.5 & 9.3 & \textbf{9.0} & 13.1 & \multicolumn{1}{r|}{12.2} \\ \hline
\multicolumn{1}{|c|}{6h runs} & $lim100$ & $lim2K$ & $lim10K$ & $lim20K$ & $ada2$ & \multicolumn{1}{r|}{$ada4$} \\ \hline
\multicolumn{1}{|r|}{0.0001 \%} & 70.1 & 82.9 & 92.2 & 97.6 & 70.5 & \multicolumn{1}{r|}{\textbf{68.8}} \\
\multicolumn{1}{|r|}{0.001 \%} & 54.6 & 61.2 & 94.5 & 96.0 & 55.1 & \multicolumn{1}{r|}{\textbf{54.3}} \\
\multicolumn{1}{|r|}{0.01 \%} & 38.6 & 37.4 & 51.1 & 81.4 & 38.0 & \multicolumn{1}{r|}{\textbf{36.3}} \\
\multicolumn{1}{|r|}{0.1 \%} & 22.8 & 22.6 & 31.4 & 21.9 & 21.9 & \multicolumn{1}{r|}{\textbf{21.1}} \\
\multicolumn{1}{|r|}{1 \%} & 12.2 & 7.9 & \textbf{7.1} & 7.2 & 7.7 & \multicolumn{1}{r|}{7.3} \\ \hline
\end{tabular} \hspace{1em}
}
\quad
\resizebox{0.45\linewidth}{!}{
\begin{tabular}{rrrrrrr}
\multicolumn{2}{l}{28-bit square} & \multicolumn{1}{l}{} & \multicolumn{1}{l}{} & \multicolumn{1}{l}{} & \multicolumn{1}{l}{} & \multicolumn{1}{l}{} \\ \hline
\multicolumn{1}{|c|}{1h runs} & $lim100$ & $lim2K$ & $lim10K$ & $lim20K$ & $ada2$ & \multicolumn{1}{r|}{$ada4$} \\ \hline
\multicolumn{1}{|r|}{0.0001 \%} & \textbf{95.0} & 97.0 & 98.4 & 98.6 & 95.8 & \multicolumn{1}{r|}{96.2} \\
\multicolumn{1}{|r|}{0.001 \%} & 90.4 & 91.0 & 96.6 & 97.9 & \textbf{90.3} & \multicolumn{1}{r|}{92.3} \\
\multicolumn{1}{|r|}{0.01 \%} & \textbf{50.6} & 81.2 & 96.4 & 97.2 & 60.2 & \multicolumn{1}{r|}{62.6} \\
\multicolumn{1}{|r|}{0.1 \%} & 30.2 & 67.2 & 95.1 & 95.8 & \textbf{19.0} & \multicolumn{1}{r|}{30.1} \\
\multicolumn{1}{|r|}{1 \%} & 9.0 & 16.3 & 9.1 & \textbf{6.6} & 9.9 & \multicolumn{1}{r|}{8.6} \\ \hline
\multicolumn{1}{|c|}{6h runs} & $lim100$ & $lim2K$ & $lim10K$ & $lim20K$ & $ada2$ & \multicolumn{1}{r|}{$ada4$} \\ \hline
\multicolumn{1}{|r|}{0.0001 \%} & \textbf{56.1} & 77.1 & 93.0 & 96.8 & 67.5 & \multicolumn{1}{r|}{70.4} \\
\multicolumn{1}{|r|}{0.001 \%} & \textbf{31.4} & 40.1 & 81.0 & 87.9 & 32.0 & \multicolumn{1}{r|}{32.3} \\
\multicolumn{1}{|r|}{0.01 \%} & 20.6 & 22.7 & 73.9 & 86.4 & 21.0 & \multicolumn{1}{r|}{\textbf{20.5}} \\
\multicolumn{1}{|r|}{0.1 \%} & 12.3 & 16.0 & 57.9 & 43.3 & 12.2 & \multicolumn{1}{r|}{\textbf{12.0}} \\
\multicolumn{1}{|r|}{1 \%} & 6.5 & 4.6 & 4.1 & \textbf{4.0} & 4.8 & \multicolumn{1}{r|}{4.2} \\ \hline
\end{tabular}
}
\end{table*}

\renewcommand{\arraystretch}{1.1}
\begin{table*}[]
\centering
\setlength{\tabcolsep}{5pt}
\caption{The overall versatility scores for the considered strategies aggregated
over the bit-widths for each of the circuits.\vspace{0.5em}}
\label{table:xfactor}
\begin{tabular}{|l|cc|cc|cc|cc|c|}
\hline
 & \multicolumn{2}{c|}{Multiplier} & \multicolumn{2}{c|}{Divider} & \multicolumn{2}{c|}{MAC} & \multicolumn{2}{c|}{Square} & \multicolumn{1}{l|}{} \\
\textbf{time limit} & 1h & 6h & 1h & 6h & 1h & 6h & 1h & 6h & \textbf{AVG} \\ \hline
$lim100$ & 104.1 & 106.9 & 121.7 & 124.1 & 109.8 & 114.4 & 116.3 & 115.4 & \textbf{114.1} \\
$lim2K$ & 113.7 & 101.3 & 113.6 & 117.2 & 102.2 & 104.5 & 160.9 & 117.7 & \textbf{116.4} \\
$lim10K$ & 201.7 & 118.8 & 102.6 & 103.0 & 126.3 & 106.0 & 196.7 & 206.6 & \textbf{145.2} \\
$lim20K$ & 322.2 & 136.0 & 101.2 & 100.8 & 151.1 & 113.2 & 193.5 & 208.2 & \textbf{165.8} \\ \hline
$ada2$ & 102.5 & 101.1 & 116.6 & 106.4 & 108.0 & 102.6 & 120.2 & 106.6 & \textbf{108.0} \\
$ada4$ & 100.6 & 100.5 & 111.9 & 104.2 & 108.8 & 101.7 & 118.7 & 103.6 & \textbf{106.3} \\ \hline
\end{tabular}
\end{table*}

In order to effectively evaluate the overall performance and versatility of the
different strategies, we introduce a~\emph{versatility score}. For each
experimental setting, we set the versatility score of the strategy $B$ that found
the best solution to $100~\%$, and other strategies $S$ are assigned the score
of $area(S) / area(B) * 100~\%$. In other words, this measure shows how many per
cent larger the solution obtained by the chosen strategy is with respect to the
best solution for the experiment (i.e.,~a~lower score is better). As before, we
compute the score from the median solutions produced by 50 independent
evolutionary runs.

Table~\ref{table:xfactor} shows the versatility scores of the inspected
strategies computed for particular circuits considering 1 and 6 hours runs.
These scores aggregate the results presented in
Tables~\ref{table:mult}--\ref{table:square} and give us a better comparison among
the strategies. The right-most column of Table~\ref{table:xfactor} contains the
versatility scores aggregated over all experiments. These scores allow us to
answer the research question $Q2$, namely, we can compare the versatility of the
fix limit strategies and the selected adaptive strategies.    

The best versatility is achieved by the adaptive strategy $ada4$. The score
$106.3$ shows that a median solution produced by this strategy is on average
about $8$~percentage points worse than a median solution produced by the best strategy for a
given experimental scenario. Strategy $ada4$ is closely followed by $ada2$ which
is by roughly $2$~percentage points worse. The best performance from
fixed-limit strategies is provided by $lim100$ that has the versatility score of $114.1$.

\new{However, since the final values are computed as averages, the final ranking is skewed by $lim2K$'s poor performance for some problem instances in square circuit approximation (see Table~\ref{table:square}: 1h runs for $0.01 \%$~WCAE). If we excluded these experiments from the final evaluation, $lim2K$ would perform considerably better than $lim100$.} 


\begin{table}[]
\caption{The pair-wise  \texttt{p-values} obtained using Nemenyi post-hoc test evaluated over all strategies and conducted experiments.}
\label{table:friedman}
\setlength{\tabcolsep}{4pt}
\centering
\renewcommand{\arraystretch}{1.1}
\begin{tabular}{|llllll|}\hline
\multicolumn{1}{|l|}{} & lim100 & lim2K & lim10K & lim20K & ada2 \\ \hline
\multicolumn{1}{|l|}{lim2K} & 0.39256 & - & - & - & - \\
\multicolumn{1}{|l|}{lim10K} & 0.99828 & 0.66932 & - & - & - \\
\multicolumn{1}{|l|}{lim20K} & 0.87598 & 0.02957 & 0.64037 & - & - \\
\multicolumn{1}{|l|}{ada2} & 0.00045 & 0.21525 & 0.00252 & 1.90E-06 & - \\
\multicolumn{1}{|l|}{ada4} & 8.00E-08 & 0.00124 & 9.30E-07 & 5.50E-11 & 0.55151\\\hline
\end{tabular}
\end{table}

\begin{figure}[]
\centering
\includegraphics[width=0.7\columnwidth]{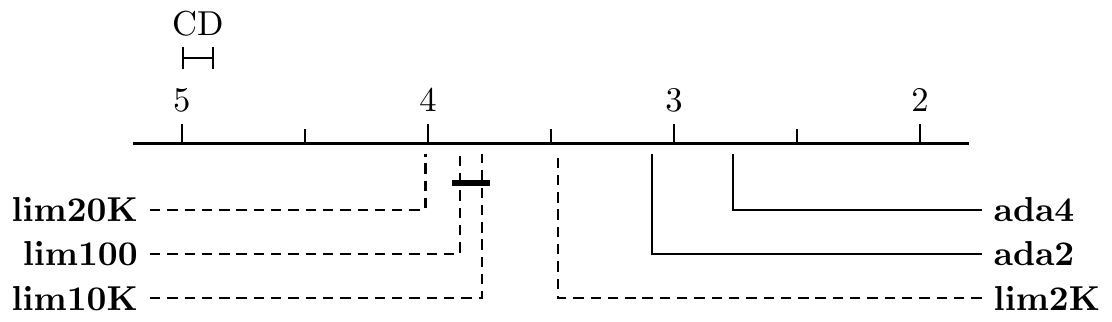}
\caption{The pair-wise comparison  of  all  strategies  obtained by Nemenyi  test.   Groups  of  strategies that  are  not significantly different (at $\texttt{p-value} = 0.05$) are connected.}
\label{fig:nemenyi}
\end{figure}

\new{We further perform Friedman statistical test with Nemenyi post hoc analysis to assess the significance of the results we obtained. In particular, we analyse the statistical significance of the versatility scores for particular approximation strategies across all conducted experiments. Friedman test returns $\texttt{chi-squared}= 71.39$ and $\texttt{p-value} < 5.2\mbox{E-14}$. These values clearly demonstrate that the  versatility scores for particular strategies are not statistically equivalent. Therefore, we use Nemenyi post hoc analysis
to identify the groups of statistically equivalent strategies. Table~\ref{table:friedman}  shows the pair-wise $p$-values for all strategy pairs. Note that these values take into consideration the evaluation  over all strategies and conducted experiments. Figure~\ref{fig:nemenyi} illustrates the average ranks (with respect to the versatility scores) of examined strategies and also visualises the groups that are not significantly different at $\alpha = 0.05$ We can conclude that strategy $ada4$ is highly significantly better ($\texttt{p-value} <0.01$) than all examined fixed limit strategies.}


\new{The statistical methods are rank based and thus they do not suffer from excessive sensibility to a few experiments with major differences in performance. Interestingly, the final placings in Fig.~\ref{fig:nemenyi} (rank based) and Table~\ref{table:xfactor} (average based) are identical with the exception of $lim100$. $Lim100$ provides decent solutions for each problem instance, hence scores well in the average based versatility score. On the other hand, it is slightly outperformed in each case by other strategies and so it's rank is even worse than that of $lim10K$. Except for a few experiment instances, $lim2K$ places among the top strategies and comes third in the rank based rankings.}

\vspace{1em}
\noindent
\new{\textbf{Summary for Q2:} The adaptive strategies, in contrast to the fixed-limit strategies, are able to provide very good performance for a wide class of approximation problems. This is demonstrated by the highest versatility score as well as by the statistical significance tests.}

\subsection{A comparison of adaptive and fixed-limit strategies (Q3)}

We saw that the adaptive strategies provide the best versatility score \new{as well as rank score} which
indicates that they can effectively handle various approximation scenarios. In
this section, we look closer at the results  presented in
Tables~\ref{table:mult}--\ref{table:square} and focus on interesting data points
revealing weak and strong properties of the adaptive strategies. In particular, we
will discuss if a single adaptive strategy can outperform the best fixed-limit
strategy for a given circuit approximation problem.

Table~\ref{table:mult} shows that the adaptive strategies dominate in almost all
approximation scenarios for multipliers. In two scenarios, the strategy $lim2K$
slightly outperforms the adaptive strategies, however, it significantly lags
behind for 1 hour runs and selected WCAEs (i.e. 24-bit version and $0.001~\%$
WCAE).    

On the other hand, the adaptive strategies lack behind the best strategies
mainly in two sets of experiments:
MACs in 1 hour evolution and dividers in 1 hour evolution (see Tables~\ref{table:mac}). Their performance is similar to that of $lim100$, and
they are outperformed by strategies with higher limit values. Since the adaptive
strategies are designed to keep the limit as low as possible while still
achieving some improvements in the candidate solutions, they do not increase
their limit value during the first hour of the experimental evaluation. Our
experiments show that even with a low resource limit, the strategies find
improvements, but many of the candidate solutions are rejected because they
cannot be evaluated within the limit. The difference in performance is
diminished as the optimisation process continues and the adaptive strategies
increase their resource limit. After 6 hours, 
the adaptive strategies outperform other settings for MACs and come close to the
performance of $lim10K$ and $lim20K$ for dividers.

\new{In case of square approximation, the adaptive strategies always produce a solution that is either the best or close to the best solution found. The exceptions are 1-hour runs for $0.01 \%$ WCAE, and 6-hour run for 28-bit version and $0.0001 \%$ WCAE, where $lim100$ significantly outperforms the other strategies.}  

\vspace{1em}
\noindent
\new{\textbf{Summary for Q3:} The adaptive strategies provides the best performance (or are very close) for a wide class of approximation problems except for MACs with the short approximation time where low-limit strategies are slightly better due to faster convergence, and for dividers where high-limit strategies are better, due to the initial phase of the adaptive~strategies.}

%
%

\subsection{A comparison with state-of-the-art techniques
(Q4)}\label{soa-comparison}

\begin{figure}[h]
\includegraphics[width=\columnwidth]{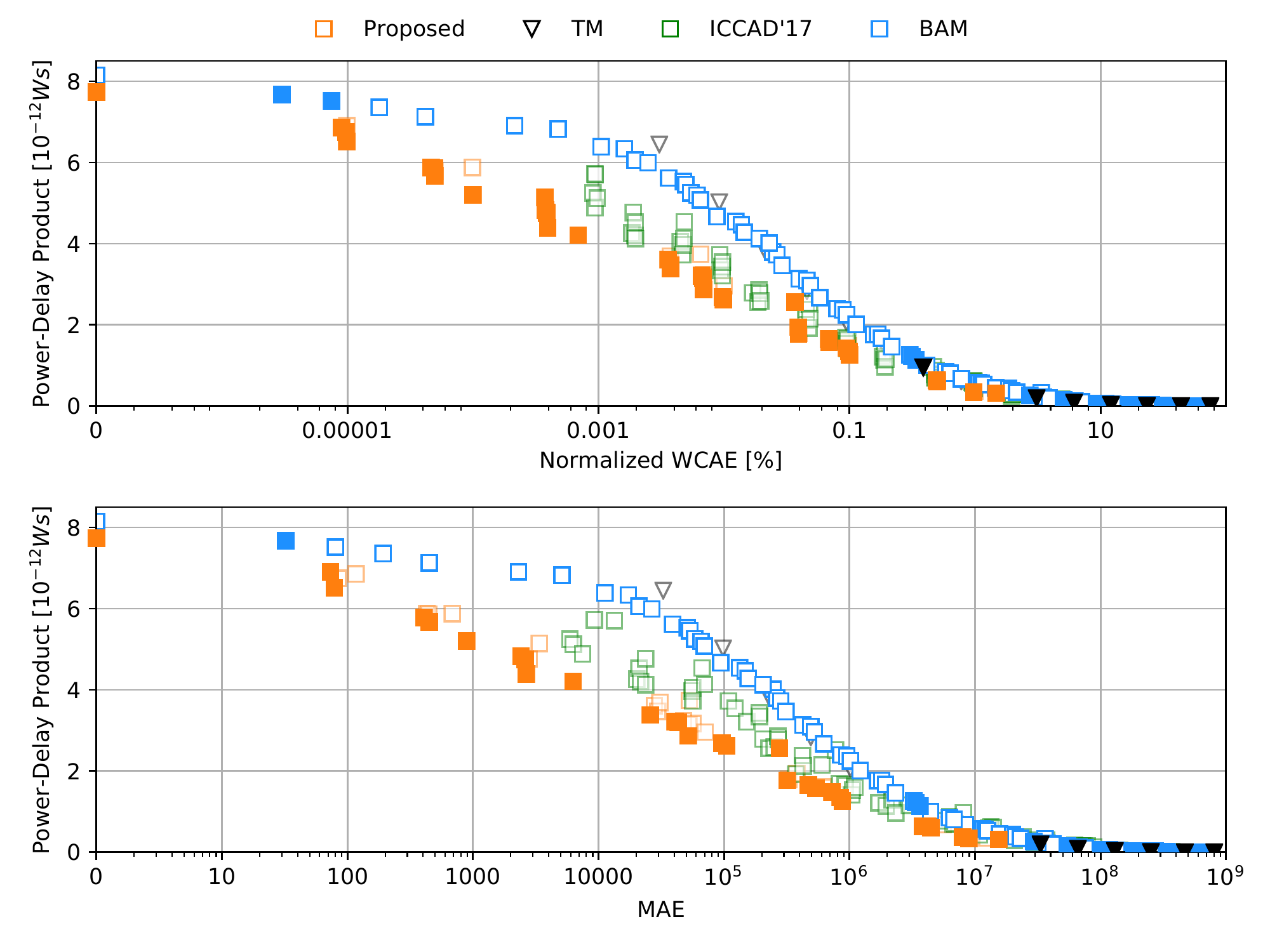}
\vspace{-2.5em}
\caption{A comparison of 16-bit approximate multipliers obtained using the
proposed approach and state-of-the-art approximation techniques. The plots show
Pareto optimal solutions and their trade-offs between the precision and the
power-delay-product (PDP)---the top plot depicts WCAE while the bottom plot
depicts the mean absolute error (MAE). The filled marks represent solutions
providing the best PDP for the given precision.}
\label{mult16-comparison}
\end{figure}

In this section, we demonstrate that our adaptive approach generates approximate circuits
that significantly outperform circuits obtained using  state-of-the-art
approximation techniques. In particular, we show that our circuits provide
significantly better trade-offs between the precision and energy consumption. We
focus on multipliers since their approximation represents a challenging and
widely studied problem---see, e.g., the comparative study of~\cite{Jiang:2015}.
On the other hand, the existing literature does not offer a sufficient number of
high-quality approximate MACs or dividers to carry out a fair comparison:
indeed, our work is the first one that automatically handles such circuits. 


\new{In the comparison, we consider two approximate architectures for multipliers
that are known to provide the best results, namely \emph{truncated multipliers}
(TMs) that ignore the values of least significant bits and \emph{broken-array
multipliers} (BAMs)~\cite{bam}. TMs and BAMs can be parameterised to
produce approximate circuits for the given  bit-width and the required error. 
In contrast to our search-based approach, these circuits are constructed using a simple deterministic procedure based on simplifying accurate multipliers. However, the method is applicable for design of approximate multipliers only. To demonstrate the practical impact of the proposed adaptive strategy, 
we also consider circuits presented in~\cite{iccad17} obtained using
verifiability-driven approximation with a fixed limit strategy --- this is a prominent representative of the search-based strategies.

Fig.~\ref{mult16-comparison} shows the parameters of resulting circuits belonging to Pareto front. For each circuit, the figure
illustrates the trade-off between the precision and the power-delay-product (PDP)
that adequately captures both the circuit's energy consumption and its delay.
The top plot of the figure illustrates the WCAE--PDP trade-offs. We also
evaluated the mean absolute error (MAE)~\cite{Error-Quant-DAC'16} of the
solutions since MAE represents another important circuit error metric. The
results are presented in the bottom plot of the figure.

The orange boxes represent circuits obtained using the adaptive strategy $ada4$. The green boxes represent circuits presented in~\cite{iccad17} and obtained using the fixed limit strategy $lim20K$. In both cases, the circuits were generated as
follows: we selected 15 target values of WCAE (10 values for the strategy $lim20K$) and
for each of these values, we executed 50 independent 2-hour runs using $\lambda=1$ and the mutation frequency~$0.5\,\%$. The 10
best solutions for each WCAE were selected and synthesised to the target technology. Note that the strategy $lim20K$ provides a much smaller reduction of the chip area when very small values of WCAE are required and thus these small target values were not reported in~\cite{iccad17}.}

\new{As we have already shown in~\cite{iccad17}, the fixed-limit verifiability-driven approach
leveraging SAT-based circuit evaluation is able to significantly outperform both
TMs and BAMs and represents state-of-the-art approximation method for arithmetic circuits.
Still, Fig.~\ref{mult16-comparison} shows that the proposed
adaptive strategy improves our previously obtained results even further---given
the same time limit, it generates circuits having significantly better
characteristics.}

\vspace{1em}
\noindent
\new{\textbf{Summary for Q4:} The proposed approach combining the SAT-based candidate evaluation with the adaptive verifiability-driven search strategy provides a fundamental improvement of the performance and versatility over existing circuit approximation techniques.}


\section{Conclusion}

Automated design of approximate circuits with formal error guarantees is a
landmark of provably-correct construction of energy-efficient systems. We
present a new approach to this problem that uniquely integrates evolutionary
circuit optimisation and SAT-based verification techniques via a novel adaptive
verifiability-driven search strategy. By being able to construct high-quality
Pareto sets of circuits including complex multipliers, MACs, and dividers, our
method shows unprecedented scalability and versatility, and paves the way for
design automation of complex approximate circuits. 

In the future, we plan to extend our approach towards different error metrics
and further classes of approximate circuits. We will also integrate the
constructed circuits into real-world energy-aware systems to demonstrate
practical impacts of our~work.

\textit{Acknowledgments:} This work was partially supported by the IT4IXS: IT4Innovations
Excellence in Science project (LQ1602) and the Brno PhD. Talent scholarship program.





\bibliographystyle{elsarticle-num.bst}
\bibliography{bibliography}

\begin{thebibliography}{10}
\expandafter\ifx\csname url\endcsname\relax
  \def\url#1{\texttt{#1}}\fi
\expandafter\ifx\csname urlprefix\endcsname\relax\def\urlprefix{URL }\fi
\expandafter\ifx\csname href\endcsname\relax
  \def\href#1#2{#2} \def\path#1{#1}\fi

\bibitem{median-filters}
Z.~Vasicek, V.~Mrazek, Trading between quality and non-functional properties of
  median filter in embedded systems, Genetic Programming and Evolvable Machines
  18~(1) (2017) 45--82.

\bibitem{Vasicek:DATE17}
Z.~Vasicek, et~al., Towards low power approximate {DCT} architecture for {HEVC}
  standard, in: Proceedings of the Conference on Design, Automation and Test in
  Europe, DATE '17, 2017.

\bibitem{bio-inspired}
H.~R. Mahdiani, et~al., Bio-inspired imprecise computational blocks for
  efficient vlsi implementation of soft-computing applications, TCAS-I (2010)
  850 -- 862.

\bibitem{Mrazek:iccad16}
V.~Mrazek, et~al., Design of power-efficient approximate multipliers for
  approximate artificial neural networks, in: Proc. of ICCAD'16, ACM, 2016, pp.
  811--817.
\newblock \href {http://dx.doi.org/10.1145/2966986.2967021}
  {\path{doi:10.1145/2966986.2967021}}.

\bibitem{vasicek:sekanina:tec}
Z.~Vasicek, L.~Sekanina, Evolutionary approach to approximate digital circuits
  design, IEEE Transactions on Evolutionary Computation 19~(3) (2015) 432--444.
\newblock \href {http://dx.doi.org/10.1109/TEVC.2014.2336175}
  {\path{doi:10.1109/TEVC.2014.2336175}}.

\bibitem{Nepal:17}
K.~Nepal, S.~Hashemi, H.~Tann, R.~I. Bahar, S.~Reda, Automated high-level
  generation of low-power approximate computing circuits, IEEE Transactions on
  Emerging Topics in Computing\href
  {http://dx.doi.org/10.1109/TETC.2016.2598283}
  {\path{doi:10.1109/TETC.2016.2598283}}.

\bibitem{mrazek:date:17}
V.~Mrazek, R.~Hrbacek, et~al., Evoapprox8b: Library of approximate adders and
  multipliers for circuit design and benchmarking of approximation methods, in:
  Proc. of DATE'17, 2017, pp. 258--261.

\bibitem{Grater:2016}
A.~Lotfi, A.~Rahimi, et~al., Grater: An approximation workflow for exploiting
  data-level parallelism in {FPGA} acceleration, in: 2016 Design, Automation
  Test in Europe Conf. Exhibition, DATE '16, EDA Consortium, 2016, pp.
  1279--1284.

\bibitem{Ciesielski16}
C.~Yu, M.~Ciesielski, Analyzing imprecise adders using {BDDs} -- a case study,
  in: Proc. of ISVLSI'16, IEEE, 2016, pp. 152--157.

\bibitem{Chand16}
A.~Chandrasekharan, M.~Soeken, D.~Große, R.~Drechsler, Approximation-aware
  rewriting of aigs for error tolerant applications, in: Proc. of ICCAD'16,
  IEEE, 2016, pp. 1 -- 8.

\bibitem{vasicek:sekanina:genp:2011}
Z.~Vasicek, L.~Sekanina, Formal verification of candidate solutions for
  post-synthesis evolutionary optimization in evolvable hardware, Genetic
  Programming and Evolvable Machines 12~(3) (2011) 305--327.

\bibitem{Ciesielski15}
M.~Ciesielski, et~al., Verification of gate-level arithmetic circuits by
  function extraction, in: Proc. of DAC '15, ACM, 2015.

\bibitem{Grobner-DATE'16}
A.~Sayed-Ahmed, D.~Gro{\ss}e, et~al., Formal verification of integer
  multipliers by combining {G}r\"{o}bner basis with logic reduction, in: Proc.
  of DATE'16, IEEE, 2016, pp. 1048--1053.

\bibitem{MACACO}
R.~Venkatesan, A.~Agarwal, K.~Roy, A.~Raghunathan, Macaco: Modeling and
  analysis of circuits for approximate computing, in: Proc. of ICCAD'11, ACM,
  2011, pp. 667--673.

\bibitem{chand:dac16}
A.~Chandrasekharan, M.~Soeken, et~al., Precise error determination of
  approximated components in sequential circuits with model checking, in: Proc.
  of DAC'16, ACM, 2016, pp. 129:1--129:6.

\bibitem{drechsler18}
S.~Frohlich, D.~Grosse, R.~Drechsler, Approximate hardware generation using
  symbolic computer algebra employing grobner basis, in: Proc. of DATE'18,
  IEEE, 2018, pp. 889--892.

\bibitem{iccad17}
M.~{\v{C}}e\v{s}ka, et~al., Approximating complex arithmetic circuits with
  formal error guarantees: 32-bit multipliers accomplished, in: Proc. of
  ICCAD'17, IEEE, 2017, pp. 416--423.

\bibitem{ADAC}
M.~{\v{C}}e{\v{s}}ka, et~al., {ADAC}: Automated design of approximate circuits,
  in: CAV'18, Vol. 10981 of LNCS, Springer, 2018.

\bibitem{tpu}
N.~P. Jouppi, C.~Young, et~al., In-datacenter performance analysis of a tensor
  processing unit, in: Proc. of ISCA'17, IEEE, 2017, pp. 1--12.

\bibitem{Mittal:2016}
S.~Mittal, A survey of techniques for approximate computing, ACM Comput. Surv.
  48~(4) (2016) 62:1--33.
\newblock \href {http://dx.doi.org/10.1145/2893356}
  {\path{doi:10.1145/2893356}}.

\bibitem{Chippa:dac2013}
V.~K. Chippa, et~al., Analysis and characterization of inherent application
  resilience for approximate computing, in: Proc. of DAC'13, ACM, 2013, pp.
  1--9.
\newblock \href {http://dx.doi.org/10.1145/2463209.2488873}
  {\path{doi:10.1145/2463209.2488873}}.

\bibitem{Xu:2016}
Q.~Xu, T.~Mytkowicz, N.~S. Kim, Approximate computing: A survey, IEEE Design
  Test 33~(1) (2016) 8--22.
\newblock \href {http://dx.doi.org/10.1109/MDAT.2015.2505723}
  {\path{doi:10.1109/MDAT.2015.2505723}}.

\bibitem{SALSA}
S.~Venkataramani, et~al., {SALSA}: systematic logic synthesis of approximate
  circuits, in: Proc. of DAC'12, ACM, 2012, pp. 796--801.

\bibitem{SASIMI:Venkataramani:date2013}
S.~Venkataramani, K.~Roy, A.~Raghunathan, Substitute-and-simplify: a unified
  design paradigm for approximate and quality configurable circuits, in: Proc.
  of DATE'13, EDA, 2013, pp. 1--6.

\bibitem{Lingamneni:2011}
A.~Lingamneni, et~al., Energy parsimonious circuit design through probabilistic
  pruning, in: Proc. of DATE'11, 2011, pp. 1--6.

\bibitem{vasicek:sekanina:ddecs2014}
Z.~Vasicek, L.~Sekanina, Evolutionary design of approximate multipliers under
  different error metrics, in: IEEE International Symposium on Design and
  Diagnostics of Electronic Circuits and Systems, IEEE, 2014, pp. 135--140.

\bibitem{Jiang:2015}
H.~Jiang, et~al., A comparative review and evaluation of approximate adders,
  in: Proc. of GLVLSI'15, ACM, 2015, pp. 343--348.

\bibitem{Li:dac2015}
C.~Li, W.~Luo, et~al., Joint precision optimization and high level synthesis
  for approximate computing, in: DAC'15, 2015, pp. 1--6.

\bibitem{Mazahir:tc16}
S.~Mazahir, O.~Hasan, et~al., Probabilistic error modeling for approximate
  adders, IEEE Trans. Comput. 66~(3) (2017) 515--530.

\bibitem{pixley04}
I.-H. Moon, C.~Pixley, Non-miter-based combinational equivalence checking by
  comparing bdds with different variable orders, in: Proc. of FMCAD'04,
  Springer, 2004, pp. 144--158.

\bibitem{abc-iprove}
A.~Mishchenko, S.~Chatterjee, R.~Brayton, N.~Een, Improvements to combinational
  equivalence checking, in: Proc. of ICCAD'06, ICCAD '06, ACM, 2006, pp.
  836--843.

\bibitem{drechsler16-grobner-cec}
A.~Sayed-Ahmed, D.~Grosse, et~al., Equivalence checking using {Grobner} bases,
  in: Proc. of FMCAD'16, IEEE, 2016, pp. 169--176.

\bibitem{Vasicek:ddecs2017}
Z.~Vasicek, Relaxed equivalence checking: a new challenge in logic synthesis,
  in: Proc. of DDECS'17, 2017, pp. 1--6.
\newblock \href {http://dx.doi.org/10.1109/DDECS.2017.7968435}
  {\path{doi:10.1109/DDECS.2017.7968435}}.

\bibitem{image}
Z.~Vasicek, et~al., Evolutionary functional approximation of circuits
  implemented into {FPGAs}, in: Proc. of SSCI'16, IEEE, 2016, pp. 1--8.

\bibitem{vasicek:so-mo}
Z.~Vas{\'i}cek, L.~Sekanina, Circuit approximation using single- and
  multi-objective cartesian gp, in: EuroGP, 2015.

\bibitem{miller:cgp:book}
J.~F. Miller, P.~Thomson, Cartesian genetic programming, in: Genetic
  Programming, Springer Berlin Heidelberg, 2000.

\bibitem{Error-Quant-DAC'16}
A.~Chandrasekharan, et~al., Precise error determination of approximated
  components in sequential circuits with model checking, in: Proc. of DAC'16,
  ACM, 2016, pp. 129:1--129:6.

\bibitem{han:cav12}
C.-S. Han, J.-H.~R. Jiang, When boolean satisfiability meets gaussian
  elimination in a simplex way, in: Proc. of CAV'12, Springer Berlin
  Heidelberg, 2012, pp. 410--426.

\bibitem{sat-backtrack}
I.~Lynce, J.~Marques-Silva, Efficient data structures for backtrack search sat
  solvers, in: Annals of Mathematics and Artificial Intelligence, Kluwer
  Academic Publishers, 2005, pp. 137--152.

\bibitem{algebraic-circuits}
A.~L. Ruiz, E.~C. Morales, L.~P. Roure, A.~G. Ríos, Algebraic circuits, in:
  Algebraic Circuits, Springer-Verlag, 2014, pp. 159--215.

\bibitem{vasicek:eurogp15}
Z.~Vasicek, Cartesian {GP} in optimization of combinational circuits with
  hundreds of inputs and thousands of gates, in: Proc. of EuroGP'15, LCNS 9025,
  Springer, 2015, pp. 139--150.

\bibitem{friedman:1940}
M.~Friedman, A comparison of alternative tests of significance for the problem
  of $m$ rankings, Ann. Math. Statist. 11 (1940) 86--92.

\bibitem{demsar:2006}
J.~Dem\v{s}ar, Statistical comparisons of classifiers over multiple data sets,
  J. Mach. Learn. Res. 7 (2006) 1--30.

\bibitem{pmcmr}
T.~Pohlert, The Pairwise Multiple Comparison of Mean Ranks Package (PMCMR), {R}
  package (2014).

\bibitem{bam}
F.~Farshchi, et~al., New approximate multiplier for low power digital signal
  processing, in: Proc. of CADS'13, IEEE, 2013, pp. 25--30.

\end{thebibliography}

%
%
%
\end{document}